\documentclass[lettersize,journal]{IEEEtran}
\usepackage{amsmath,amsfonts}
\usepackage{algorithmic}
\usepackage{algorithm}
\usepackage{array}
\usepackage[caption=false,font=normalsize,labelfont=sf,textfont=sf]{subfig}
\usepackage{textcomp}
\usepackage{stfloats}
\usepackage{url}
\usepackage{verbatim}
\usepackage{graphicx}
\usepackage{cite}
\usepackage{booktabs}
\usepackage{multirow}
\usepackage{adjustbox}
\usepackage{gensymb}
\usepackage{hyperref}
\usepackage{paralist}

\begin{document}

\title{SurgNavAR: An Augmented Reality Surgical Navigation Framework for Optical See-Through Head Mounted Displays}

\author{Abdullah Thabit, Mohamed Benmahdjoub, Rafiuddin Jinabade, Hizirwan S. Salim, Marie-Lise C. van Veelen, Mark G. van Vledder, Eppo B. Wolvius and Theo van Walsum, \IEEEmembership{Member, IEEE}
\thanks{This paragraph of the first footnote will contain the date on which
you submitted your paper for review.}
\thanks{Abdullah Thabit and Mohamed Benmahdjoub are with the Biomedical Imaging Group Rotterdam, Department of Radiology \& Nuclear Medicine, and also with the Department of Oral and Maxillofacial Surgery, Erasmus MC, 3015 GD Rotterdam, The Netherlands (e-mail:a.thabit@erasmusmc.nl;m.benmahjdoub@erasmusmc.nl).}
\thanks{Rafiuddin Jinabade was with the University of Groningen, 9712 CP Groningen, The Netherlands, and with the Department of Oral and Maxillofacial Surgery, Erasmus MC, 3015 GD Rotterdam, The Netherlands}
\thanks{Hizirwan S. Salim is with the Department of High Performance Compute \& Visualization, SURF bv, 3511 EP Utrecht, The Netherlands}
\thanks{Marie-Lise C. van Veelen is with the Department of Neurosurgery, Erasmus MC, 3015 GD Rotterdam, The Netherlands}
\thanks{Mark G. van Vledder is with the Trauma Research Unit, Department of Surgery, Erasmus MC, 3015 GD Rotterdam, The Netherlands}
\thanks{Eppo B. Wolvius is with the Department of Oral and Maxillofacial Surgery, Erasmus MC, 3015 GD Rotterdam, The Netherlands}
\thanks{Theo van Walsum is with the Biomedical Imaging Group Rotterdam, Department of Radiology \& Nuclear Medicine, Erasmus MC, 3015 GD Rotterdam, The Netherlands}}

\markboth{Journal of \LaTeX\ Class Files,~Vol.~14, No.~8, August~2021}%
{Shell \MakeLowercase{\textit{et al.}}: A Sample Article Using IEEEtran.cls for IEEE Journals}

\IEEEpubid{0000--0000/00\$00.00~\copyright~2021 IEEE}

\maketitle

\begin{abstract}
Augmented reality (AR) devices with head mounted displays (HMDs) facilitate the direct superimposition of 3D preoperative imaging data onto the patient during surgery. To use an HMD-AR device as a stand-alone surgical navigation system, the device should be able to locate the patient and surgical instruments, align preoperative imaging data with the patient, and visualize navigation data in real time during surgery. Whereas some of the technologies required for this are known, integration in such devices is cumbersome and requires specific knowledge and expertise, hampering scientific progress in this field. This work therefore aims to present and evaluate an integrated HMD-based AR surgical navigation framework that is adaptable to diverse surgical applications. The framework tracks 2D patterns as reference markers attached to the patient and surgical instruments. It allows for the calibration of surgical tools using pivot and reference-based calibration techniques. It enables image-to-patient registration using point-based matching and manual positioning. The integrated functionalities of the framework are evaluated on two HMD devices, the HoloLens 2 and Magic Leap 2, with two surgical use cases being evaluated in a phantom setup: AR-guided needle insertion and rib fracture localization. The framework was able to achieve a mean tooltip calibration accuracy of 1 mm, a registration accuracy of 3 mm, and a targeting accuracy below 5 mm on the two surgical use cases. The framework presents an easy-to-use configurable tool for HMD-based AR surgical navigation, which can be extended and adapted to many surgical applications. The framework is publicly available at \href{https://github.com/abdullahthabit/SurgNavAR}{https://github.com/abdullahthabit/SurgNavAR}.
\end{abstract}


\section{Introduction}
\label{sec:introduction}
\IEEEPARstart{S}{urgical} navigation has become standard in clinical care by offering surgeons visual guidance during operations and allowing safer and less invasive surgeries. Especially complex surgeries benefit from navigation for precise targeting and localization of anatomical structures~\cite{mezger2013navigation}.

Conventional navigation systems typically track the patient and surgical instruments and show the preoperative data on a 2D screen next to the operating table. Such navigation systems have high tracking accuracy, but they suffer from a few drawbacks such as presenting the 3D navigating data on 2D screens, forcing surgeons to constantly divert their attention between the patient and the display and requiring a difficult hand-eye coordination in translating the presented data to the operative field. \cite{mezger2013navigation,leger2017quantifying}. Moreover, the learning curve and ergonomic factors associated with using conventional navigation systems further hinder their integration in many surgical procedures \cite{rivkin2009challenges}.

Augmented Reality (AR) may enhance the surgeon’s vision by overlaying 3D navigation data and preoperative images directly onto the patient. This can improve hand-eye coordination and visualization of imaging data. Modern optical see-through (OST) head-mounted displays (HMDs) such as the HoloLens 2 (HL2) (Microsoft Corporation, USA) and Magic Leap 2 (ML2) (Magic Leap, USA) are compact, sensor-equipped headsets with built-in computing, making them well-suited for surgical navigation. Their use has been explored across various specialties, including orthopedic surgery~\cite{muller2020augmented,liebmann2019pedicle}, neurosurgery~\cite{ivan2021augmented,andereggen2024mixed}, and cranio-maxillofacial surgery~\cite{benmahdjoub2021augmented,doughty2022augmenting}.

\IEEEpubidadjcol

Over the recent years, a substantial number of studies have explored the use of HMD-based AR systems for surgical navigation. A wide range of prototypes and frameworks have been proposed across various surgical applications. For instance, Liebmann et al. developed an HL2-based navigation framework for surface digitization and registration to assist in pedicle screw navigation~\cite{liebmann2019pedicle}, while Martin-Gomez et al. connected the HL2 to a workstation for surgical tools tracking with reflective markers~\cite{martin2023sttar}. Similarly, Iqbal et al.combined the HL2 with a robotic platform for computer-assisted orthopedic surgery using reflective marker tracking~\cite{iqbal2022semi}, and Gsaxner et al. proposed on-device surgical tool tracking with an HL2 equipped with an additional light source~\cite{gsaxner2021inside}. In contrast, Benmahdjoub et al. introduced a hybrid system integrating the HL2 with an external tracking system~\cite{benmahdjoub2022multimodal}, whereas Li et al. implemented a marker-less registration approach to guide external ventricular drain (EVD) catheter placement~\cite{li2024evd}.

Collectively, these studies demonstrate the strong potential of HMD-based AR systems in enhancing surgical navigation and intraoperative guidance. However, most existing solutions remain limited in scope, often tailored to a specific surgical application or HMD device, and frequently dependent on additional hardware configurations beyond the AR headset itself. Furthermore, to the best of our knowledge, no open-source software currently integrates all essential components of the surgical navigation workflow while remaining adaptable to different surgical contexts. Existing studies and tools typically provide only partial functionality, such as tracking or registration, and require further development, making them inaccessible to less technical users.This inaccessibility, in return, limits the reproducibility, extensibility, and clinical validation of such tools. This situation contrasts with other areas of medical imaging, where publicly available tools have driven significant progress. For example, medical image segmentation has advanced substantially with open-source methods such as TotalSegmentator \cite{wasserthal2023totalsegmentator}, MedSAM \cite{ma2024segment}, and nnU-Net \cite{isensee2021nnu}. AR navigation research, by comparison, remains hindered by the absence of comprehensive frameworks, forcing investigators to reimplement existing methods for validation and adaptation. To accelerate progress in this field, open-source, user-friendly, and application-agnostic surgical navigation frameworks for HMD devices are needed.

This work has three main objectives: First, to introduce an open-source, application-agnostic and device-agnostic, end-to-end HMD-based AR framework for surgical navigation, designed for adaptability by less technical users. Second, to benchmark the framework’s performance and analyze its key functionalities. Third, to demonstrate and evaluate the system’s usability through surgical use cases.

\section{Background \& Related Work}
Navigation systems provide image-guided assistance during surgery, supporting tasks such as tumor localization and needle insertion. To achieve this, they must continuously track the patient and surgical instruments,register preoperative images with the patient’s anatomy, and visualize navigation data in real time. The following sections discuss these core functionalities in detail, and outline how they have been adopted for HMD-based AR surgical navigation:

\subsubsection{Tracking of patient and surgical instruments} 
To track the patient, surgical instruments, and their relative poses, a marker-based or marker-less approach can be used. In marker-based tracking, reference markers are attached to the patient and surgical tools, allowing the navigation system to detect and determine their spatial positions and orientations. Common tracking methods include electromagnetic (EM) and optical (OP) tracking \cite{sorriento2019optical}. In EM tracking, small wired coils serve as reference sensors that are detected by the system within a generated EM field. In OP tracking, an infrared camera tracks light-emitting diodes or retro-reflective spheres mounted on a rigid body. Alternatively, standard RGB cameras can be used to track 2D patterns. Software solutions such as Vuforia \cite{Vuforia}, ArUco \cite{garrido2014automatic}, and AprilTag \cite{olson2011apriltag} are commonly used to detect and estimate the 3D pose of such 2D markers.

In marker-less tracking, the patient and surgical tools are directly tracked by the system, eliminating the need for physical markers. To achieve this, computer vision algorithms are employed to detect and identify target objects within the image frames of RGB and depth cameras. Anatomical features such as facial landmarks and surface topology can be exploited to automatically localize the patient~\cite{liu2020automatic,benmahdjoub2023fiducial}. Similarly, geometric characteristics and contours of surgical instruments can be recognized from camera inputs to determine their spatial pose within the operative field~\cite{hein2025next,doughty2022hmd}.

Recent AR HMD devices such as the HL2 integrate multiple sensors including a photo-video camera, stereoscopic spatial mapping cameras, and an infrared depth camera~\cite{ungureanu2020hololens}, allowing for a variety of tracking approaches to be investigated~\cite{kunz2021autonomous, thabit2024augmented, ivanov2021intraoperative, martin2023sttar, iqbal2022semi, liebmann2019pedicle, gsaxner2021inside}. Among the proposed HMD-based AR systems employing marker-based tracking, Liebmann et al. tracked ApriTags using the stereo cameras~\cite{liebmann2019pedicle}, whereas Benmahdjoub et al. used the RGB camera to track Vuforia markers~\cite{benmahdjoub2022multimodal}. Martin-Gomez et al., Iqbal et al. and Gsaxner et al. instead tracked reflective markers rather than optical patterns~\cite{martin2023sttar,iqbal2022semi,gsaxner2021inside}. For marker-less tracking, Li et al. utilized the depth camera to localize the patient; however, they reported the need to revert to marker-based tracking for subsequent patient pose updates for improved tracking accuracy and efficiency~\cite{li2024evd}.

Although marker-less tracking minimizes disruptions to the surgical workflow, it is inherently more sensitive to intraoperative variations such as occlusions and topology changes caused by surgical draping, bleeding and other factors. Furthermore, current marker-less approaches generally support only initial localization rather than continuous real-time tracking due to their high computational demands. In contrast, marker-based tracking enables greater adaptability and generalizability of navigation systems across a wide range of surgical applications, as markers can be strategically attached to various anatomical sites and instruments. Conventional navigation systems, including the Brainlab Kick® (Brainlab AG, Germany), Stryker NAV3i® Platform (Stryker, Portage, MI, USA), Medtronic StealthStation® (Medtronic, Minneapolis, MN, USA), and ClaroNav Navient® (ClaroNav, Toronto, ON, Canada) all employ marker-based tracking achieve robust, accurate, and generalizable patient and instrument tracking.

\subsubsection{Calibration of surgical instruments}
Accurately determining the position and orientation of surgical instruments is crucial in image-guided surgery, as it enables tasks such as anatomical landmark annotation, image-to-patient alignment, guided drilling, and needle insertion~\cite{chen2020external}. The pose of a tool with respect to its attached reference marker can be predefined by design~\cite{liebmann2019pedicle} or determined through calibration~\cite{kim1999universal, surya2024design}. Common calibration methods include pivoting and template-based approaches~\cite{chen2020external}. In clinical practice, it is particularly important to enable the calibration of various instruments, such as surgical drills~\cite{kim1999universal}. Conventional navigation systems allow for surgical tool calibration at runtime. However, most proposed HMD-based AR navigation frameworks either rely on the calibration capabilities of the integrated external tracking system~\cite{benmahdjoub2022multimodal,iqbal2022semi} or assume the tool pose to be known by design~\cite{liebmann2019pedicle,gsaxner2021inside,martin2023sttar}.

\subsubsection{Image-to-patient alignment}
Image-to-patient alignment enables the transformation of preoperative image data into the patient’s coordinate system, which is required for accurate intraoperative visualization and direct overlay during surgery. This step is essential for surgical navigation. Several methods have been developed for image-to-patient alignment in computer-assisted surgery. A tracked stylus can be used to annotate anatomical landmarks, which are then matched  to corresponding points in the preoperative model using paired-points registration \cite{arun1987least, thabit2022augmented, li2019mixed}, or to digitize the patient’s surface for alignment with the model using iterative closest point (ICP) \cite{besl1992method, liebmann2019pedicle}. 

Alternatively, a depth camera can be leveraged in marker-less registration to automatically construct a comprehensive intraoperative point cloud of the patient, eliminating the need for manual surface digitization~\cite{haxthausen2021superimposing, groenenberg2024feasibility, kerkhof2025depth}. Similarly, computer vision techniques can be used to automatically detect anatomical features and align the preoperative image with the patient~\cite{benmahdjoub2023fiducial,gsaxner2021augmented}.

\begin{figure}[]
\centerline{\includegraphics[width=\columnwidth]{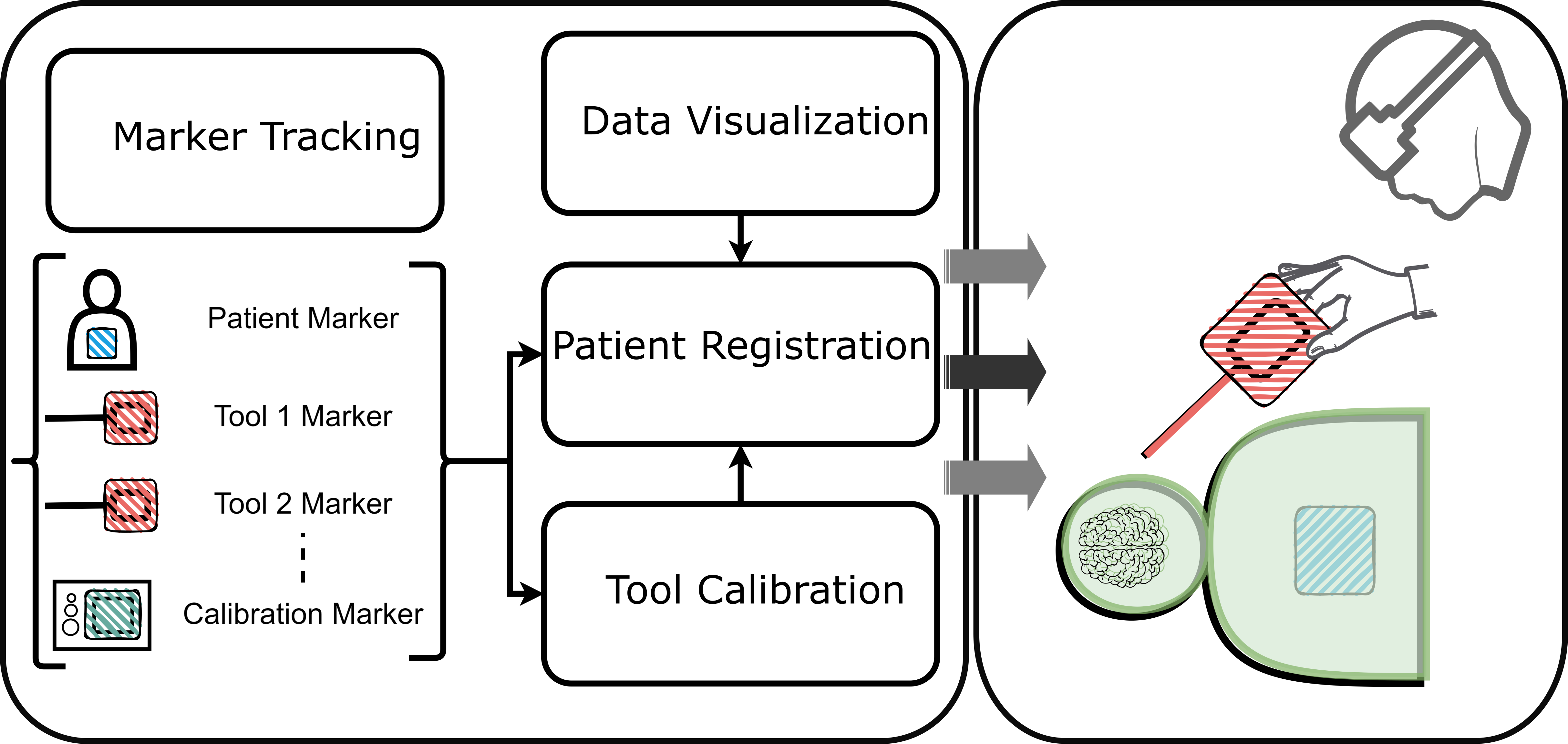}}
\caption{Overview of the framework and its navigation modules (left). The modules work together to provide AR guidance for the surgeon (right).}
\label{fig:framework_overview}
\end{figure}

The above mentioned techniques can be used with both conventional navigation systems and HMD-based AR navigation. However,  in AR HMDs, the inherent ability to visualize and manipulate  3D models  also enables the alignment of the preoperative model with the patient through manual placement \cite{pratt2018through, li2018wearable, mitsuno2019effective}. Among the proposed HMD-based systems in the literature, Benmahdjoub et al. employed point-based matching~\cite{benmahdjoub2022multimodal}, Liebman et al. and Iqbal et al. used surface registration~\cite{liebmann2019pedicle,iqbal2022semi}, while Martin-Gomez et al. performed registration using a calibration model~\cite{martin2023sttar}, and Li et al. implemented marker-less, depth-based registration to align the preoperative scan with the patient’s head~\cite{li2024evd}. Conventional navigation systems, by contrast, typically rely on point-based and surface-matching methods to achieve robust and accurate image-to-patient registration~\cite{mezger2013navigation,cleary2010image}.

\subsubsection{Preoperative and intraoperative data visualization}
A major advantage of OST AR-HMD devices in surgical navigation is their ability to display preoperative imaging data in 3D, directly within the surgeon's field of view. Various visualization methods, such as volume rendering and 3D surface rendering, can be used \cite{gehrsitz2021cinematic, pratt2018through}. The latter is often preferred for AR-HMDs due to their limited computational power and the added visual customization it offers \cite{doughty2022augmenting}.

\section{Methodology}
This section presents the proposed framework, including its requirements, functionalities, and design, followed by a detailed discussion of its modules.

\subsection{Framework Requirements and design}

To enable stand-alone HMD-based AR surgical navigation, the following is required: \begin{inparaenum}[(I)] 
    \item an HMD headset and a number of reference markers, 
    \item methods for tracking the patient and multiple (surgical) tools, 
    \item methods for tool-tip calibration of (surgical) tools.
    \item methods for image-to-patient alignment,
    \item methods for visualization.
\end{inparaenum} The proposed framework incorporates these requirements for HMD-based surgical navigation, and is made generic, device agnostic (works with two common HMD devices) and application agnostic, such that it can be easily configured for various surgical applications.

Figure \ref{fig:framework_overview} illustrates the key components of the framework and their interactions for end-to-end surgical navigation. These modules are further detailed in sections~\ref{subsec:markerTracking} - \ref{subsec:visPreData}.

\begin{figure}[]
\centerline{\includegraphics[width=0.9\columnwidth]{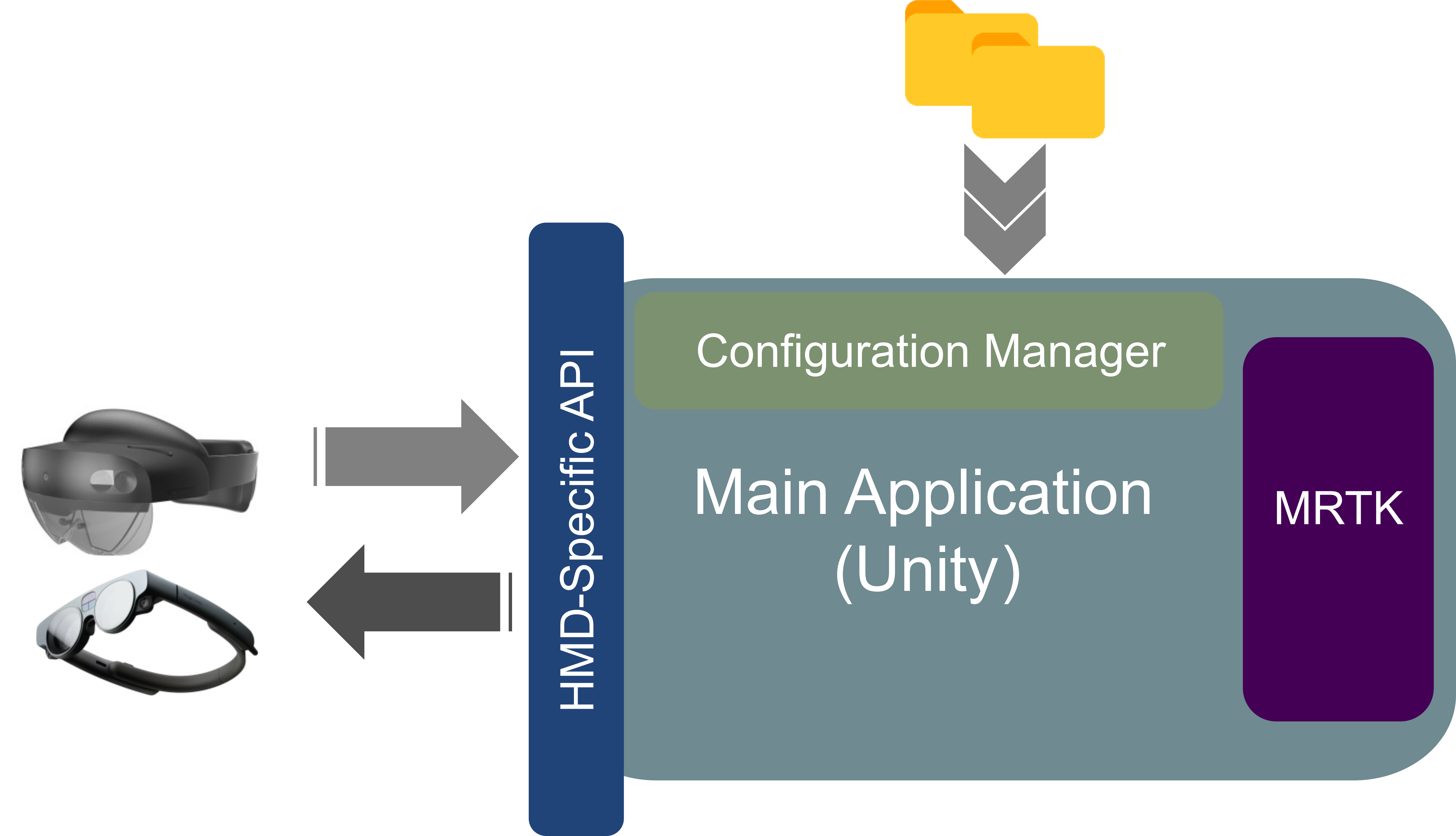}}
\caption{Framework architecture.}
\label{fig:architecture_framework}
\end{figure}

The overall architecture is shown in Figure \ref{fig:architecture_framework}. The main application integrates the navigation modules and communicates with the HMD device via an HMD-specific API. Below is a brief description of the core architectural elements:

\subsubsection{Main Application}
It is responsible for integrating the modules shown in Figure \ref{fig:framework_overview}. The main application manages the tracking data, processing algorithms and the visualization of preoperative data. It is built on Unity and uses the mixed reality toolkit (MRTK3) for creating and managing the user interface.

\subsubsection{HMD-Specific API}
It is an integration layer that connects the device's API with the main application. It accesses the HMD sensors and facilitates the communication of its data to the main application.

\subsubsection{Configuration Manager}
It controls parameters specific to the target surgical application. It reads a configuration file (stored locally in the HMD device) specifying the required functionalities and communicates them to the main application.

\subsection{Tracking of reference markers}
\label{subsec:markerTracking}
\begin{figure}[]
\centerline{\includegraphics[width=\columnwidth]{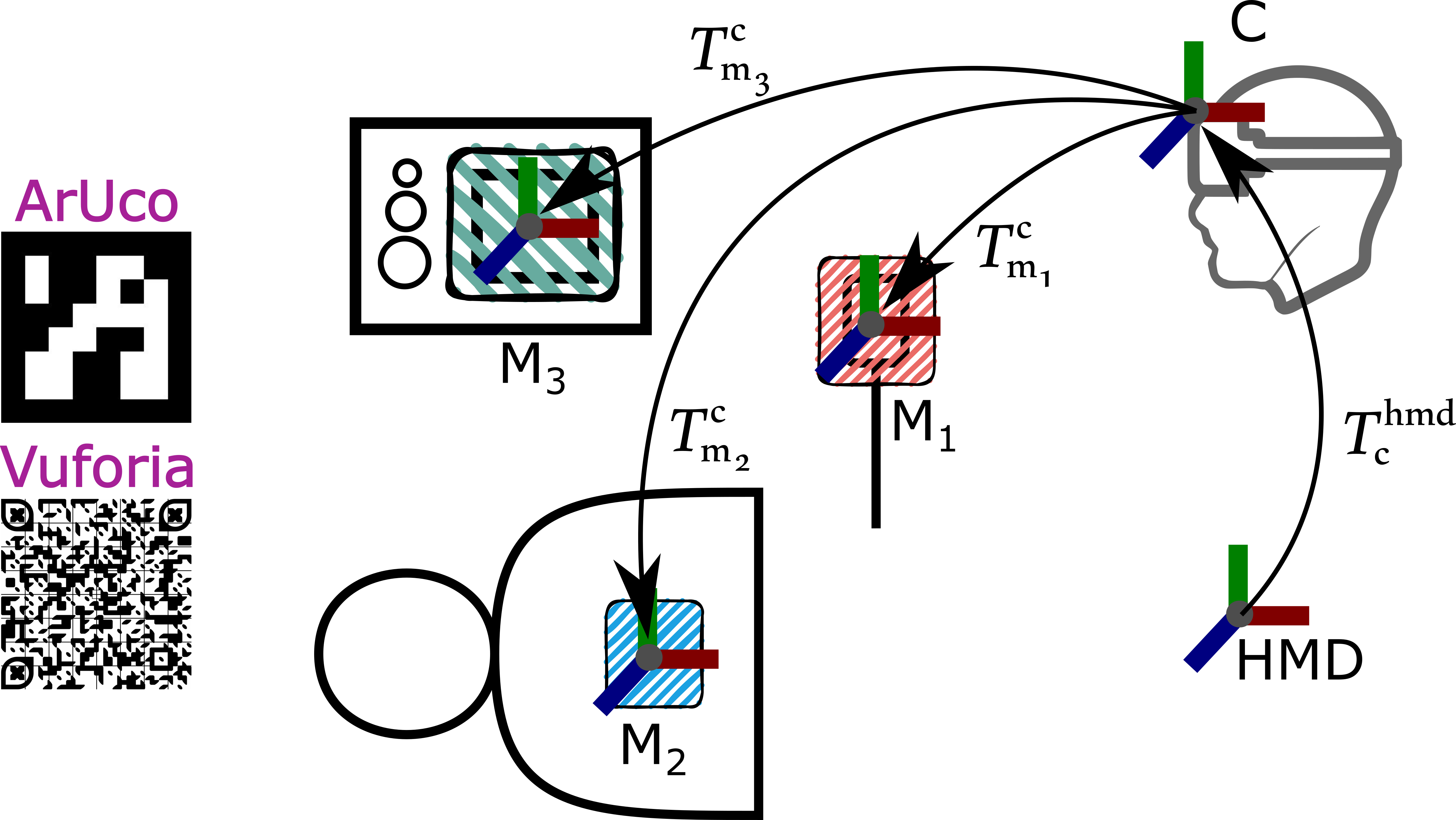}}
\caption{Tracking of reference markers in the framework with sample ArUco/AprilTag and Vuforia markers}
\label{fig:marker_tracking}
\end{figure}

Marker tracking refers to the continuous pose estimation of a marker in the camera coordinate system of the HMD device. Given that AR-HMD devices are equipped with at least one RGB camera, the 3D pose of a marker $m_i$ in the camera coordinate system $T^c_{m_i}$ can be found by detecting salient features of the marker in the image plane and mapping those feature-points to estimate the marker pose using perspective-n-point (PnP) \cite{li2012robust}.
The marker pose in the AR-HMD space $T^\mathrm{hmd}_{m_i}$ can then be determined by:

\begin{equation}
T^\mathrm{hmd}_{m_i} = T^\mathrm{hmd}_c~.~T^c_{m_i} \quad , 
\end{equation}
where $T^\mathrm{hmd}_c$ is the pose of the camera in the HMD world coordinate system (see Figure \ref{fig:marker_tracking}).

The framework supports tracking multiple markers, configurable via the configuration file. It offers two modes: tracked, when the marker is in frame, and extended-tracked, which uses the HMD's simultaneous localization and mapping (SLAM) capabilities to maintain the last known marker pose when out of frame. 

ArUco and Vuforia are widely used for 2D marker tracking in HMD-based surgical navigation and have been evaluated in various surgical applications \cite{doughty2022augmenting, kunz2021autonomous, ivanov2021intraoperative, hu2021head, ackermann2021augmented}. Both are integrated into the framework, and their tracking accuracy is assessed in the context of HMD-based AR surgical navigation. ArUco/AprilTag is implemented in the HMD-specific API and configured for each HMD device, with marker dictionaries defined at runtime. Vuforia is implemented in the main application using the Vuforia-Unity SDK, with custom image targets defined at runtime.

\subsection{Calibration of surgical instruments}
\label{subsec: toolsCalib}

In tool calibration, the tool-tip position and orientation relative to the marker attached to the tool are estimated. For a marker $m_i$ on a surgical tool, the tool-tip pose in marker coordinates is represented as $T^{m_i}_{tp_i}$ (see Figure \ref{fig:tool_calibration}a). The tool-tip pose in world coordinates is then calculated using:

\begin{equation}
T^\mathrm{hmd}_{tp_i} = T^\mathrm{hmd}_{m_i}~.~T^{m_i}_{tp_i}
\end{equation}

There are several methods to locate the tool-tip in marker coordinates $T^{m_i}_{tp_i}$. If the marker is integrated into the tool's design, the tool-tip pose is known from the tool's geometry. Otherwise, calibration is required. The framework includes three techniques for tool-tip pose estimation: pivot calibration, using a calibration tool, and marker-to-marker user calibration, which are explained below.

\subsubsection{Pivot calibration}
In this approach, the tool is pivoted around its tip at a fixed position. The tool-tip position in marker coordinates $p_m$ is then estimated by solving a system of linear equations, one for each time point. Each equation ensures that the tip position, transformed to world coordinates using the tool's marker pose $T^\mathrm{hmd}_{m_i}$, results in the same world coordinate position:

\begin{equation} \label{eq:pivot}
p_w = [R_i]~.~p_m + t_i \quad , 
\end{equation}
where $p_w$ is the tool-tip position in HMD world space. $R_i$ and $t_i$ are the rotation and translation components of the marker pose $T^\mathrm{hmd}_{m_i}$ respectively (see Figure \ref{fig:tool_calibration}b). Note that pivot calibration only estimates the position of the tool-tip, and not the orientation of the tool.

\subsubsection{Using a calibration tool}
\begin{figure}[]
\centerline{\includegraphics[width=0.8\columnwidth]{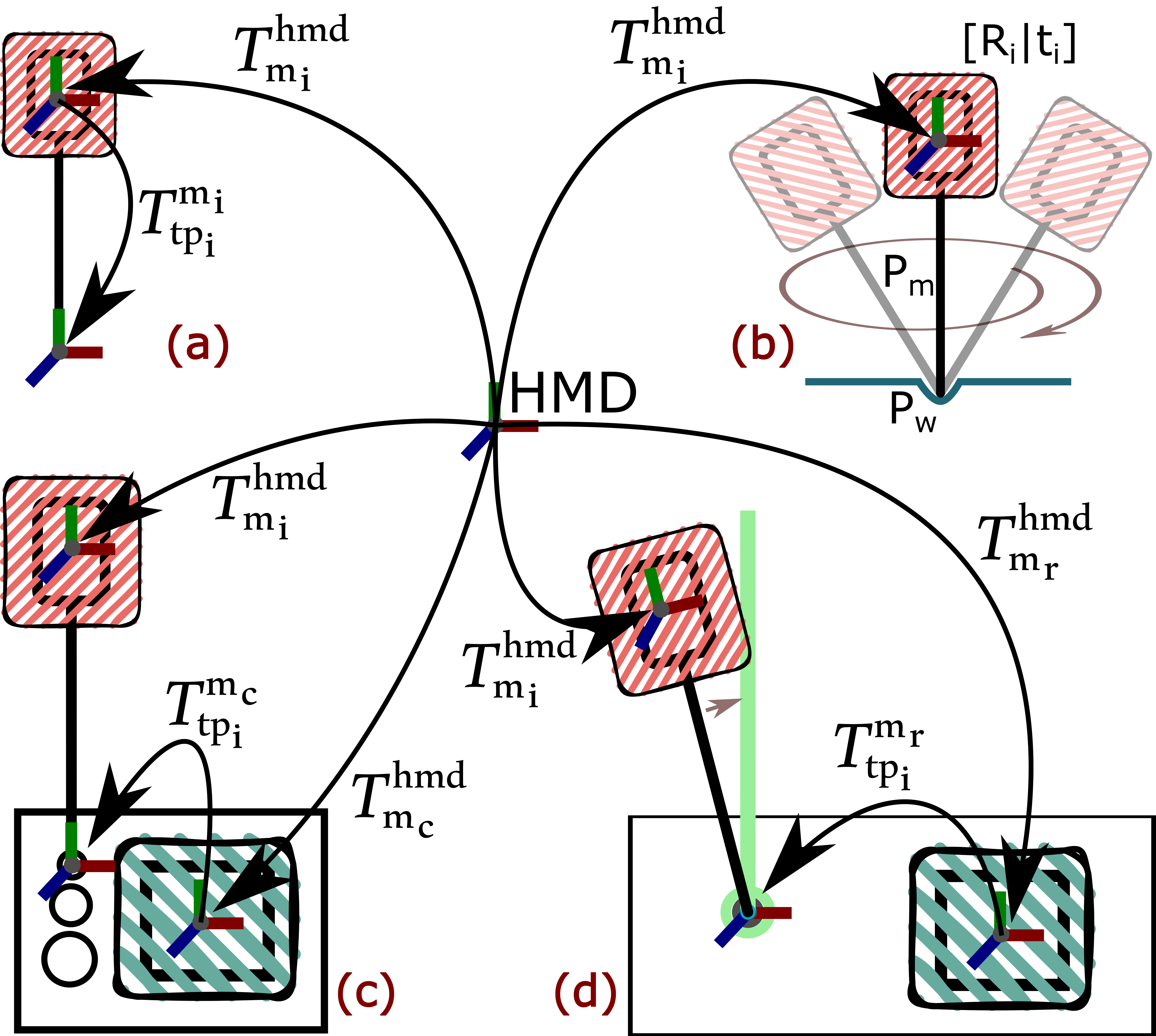}}
\caption{Calibration of surgical instruments: tool-tip pose estimation (a) using pivot calibration (b), a calibration tool (c), and marker-to-marker user calibration (d)}
\label{fig:tool_calibration}
\end{figure}

To estimate the tool-tip position and orientation relative to the tool's marker, a calibration tool can be used. The calibration tool or calibrator can have an insertion hole with the same diameter as the tool to be calibrated, and is positioned at a known distance from the calibrator's marker. When the tool to be calibrated is inserted, its pose $T^{m_i}_{tp_i}$ can be determined as follows:

\begin{equation} \label{eq:tool}
T^{m_i}_{tp_i} = (T^\mathrm{hmd}_{m_i})^{-1}~.~T^\mathrm{hmd}_{m_c} .~T^{m_c}_{tp_i}
\quad ,
\end{equation}
where $T^\mathrm{hmd}_{m_c}$ and $T^{m_c}_{tp_i}$ are the pose of the calibrator's marker in HMD space and calibration hole in calibrator's space respectively (see Figure \ref{fig:tool_calibration}c).

\subsubsection{Marker-to-marker user calibration}
This approach uses a simple calibration setup without requiring an additional calibration tool. A divot-point is printed on paper at a known position relative to a reference static marker $m_r$. A virtual cylinder is visualized at the divot-point, perpendicular to the marker. The user places the tool's tip on the divot and aligns its shaft with the virtual cylinder. The tool-tip pose is then estimated as follows:

\begin{equation} \label{eq:user-calibration}
T^{m_i}_{tp_i} = (T^\mathrm{hmd}_{m_i})^{-1}~.~T^\mathrm{hmd}_{m_r} .~T^{m_r}_{tp_i}
\quad ,
\end{equation}
where $T^\mathrm{hmd}_{m_r}$ and $T^{m_r}_{tp_r}$ are the pose of the static marker in HMD space and virtual cylinder in static marker's space respectively (see Figure \ref{fig:tool_calibration}d).

\subsection{Image-to-patient alignment}
To determine the transformation needed to bring the preoperative data from the image space to the HMD space $T^\mathrm{hmd}_\mathrm{img}$, two methods are implemented in the framework: manual-positioning and point-based matching.

\subsubsection{Manual-positioning}
Given that AR-HMD devices enable intuitive object manipulation using hand tracking and tracked controllers, manual positioning can be used for image-to-patient alignment. In the framework, virtual models can be moved in 6 degrees of freedom (DOF) or in individual 1-DOF for precise alignment, enabling the direct overlay of virtual models on the target anatomy.

\subsubsection{Point-based matching}
\begin{figure}[]
\centerline{\includegraphics[width=0.7\columnwidth]{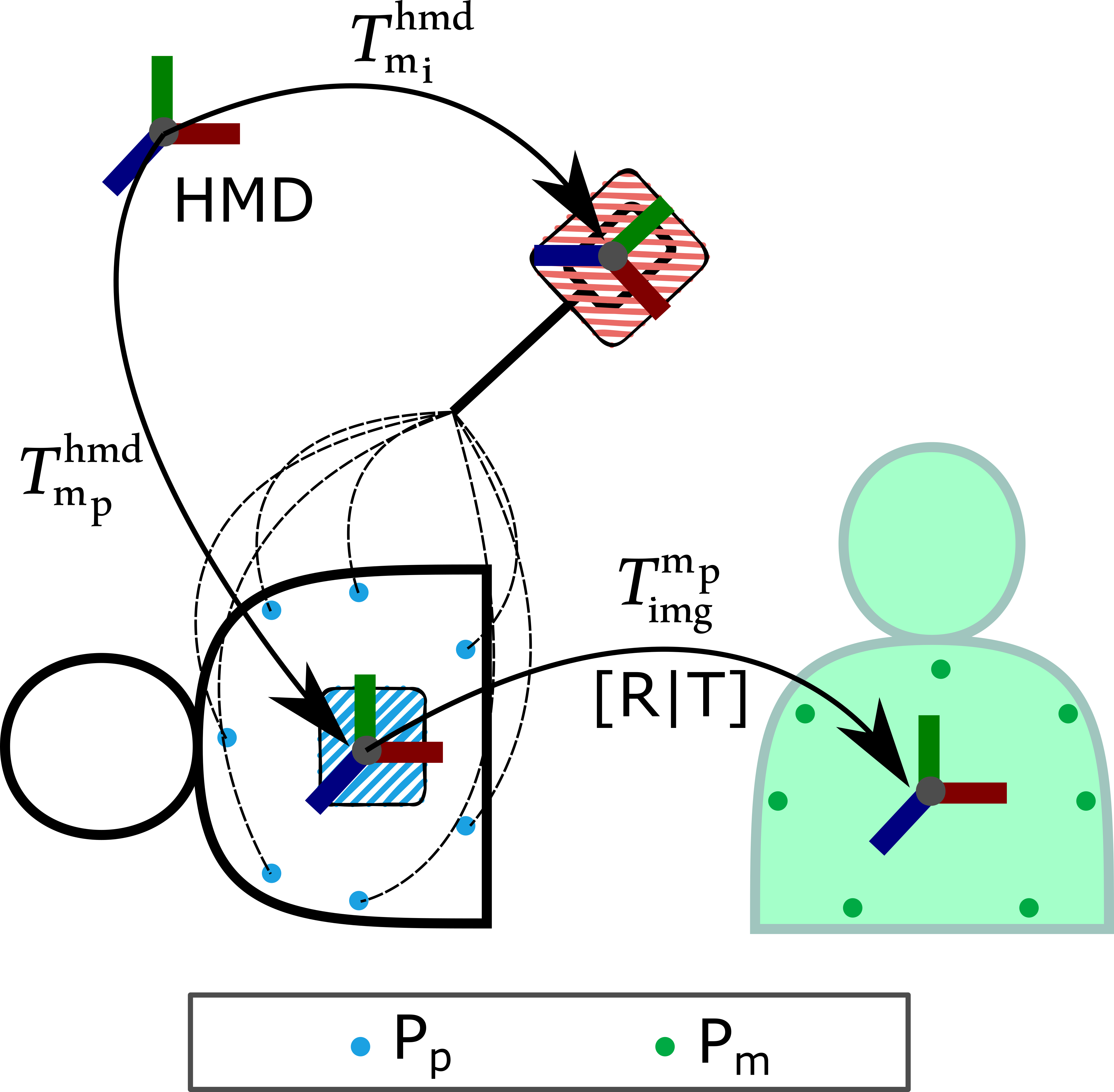}}
\caption{Image-to-patient alignment using point-based matching between the virtual model and patient.}
\label{fig:image_registration}
\end{figure}

In point-based matching, anatomical and fiducial landmarks are annotated on the patient with a tracked pointer and matched to their counterparts on the preoperative model to determine the image pose in HMD space $T^\mathrm{hmd}_\mathrm{img}$ (see Figure \ref{fig:image_registration}). This method offers a more direct, less user-dependent alignment, compared to manual positioning. Given two sets of points, model points $P_m$ and patient points $P_p$, they are matched using:

\begin{equation}
    P_p = R~.~P_m + T \quad ,
\end{equation}
where $R$ and $T$ are the rotation and translation components of the registration matrix, determined by minimizing the mean squared error through singular value decomposition \cite{arun1987least}.

\subsection{Data visualization}
\label{subsec:visPreData}
To visualize preoperative data on the HMD device, the framework uses surface mesh visualization instead of volumetric rendering, due to HMD devices' limited computational power. Target structures are segmented from CT or MRI images to generate polygon models, which can then be loaded at runtime. Multiple target structures can be visualized simultaneously, with the user controlling visibility, opacity, material, and color for optimal display. Additionally, preoperative planning annotations, such as landmark and trajectory annotations (based on entry and exit points), can be visualized for intraoperative guidance.

\section{Experiments And Results}
Extensive experiments were conducted to evaluate each of the modules, along with an end-to-end assessment for two clinical applications. The framework was adapted and tested on two state-of-the-art AR HMD devices: the Microsoft HoloLens 2 (HL2) and Magic Leap 2 (ML2).

An NDI Vega optical tracking system (OTS) was used as a reference to assess the framework’s accuracy across various surgical tasks. The OTS tracked two passive 4-marker rigid bodies and a pointer, with its role explained for each experiment separately. For the setup, two Vuforia image targets were created and tested for optimal detectability by the Vuforia SDK (five stars). These targets were 3D printed on rigid plastic plates with edge lengths of 8 cm and 5 cm, featuring 36 divots for calibration with the NDI system. Additionally, two 4x4 ArUco/AprilTag markers from the AprilTag25 family were selected. These were also 3D printed on plastic plates with edge lengths of 8 cm and 5 cm.

\begin{figure}[]
\centerline{\includegraphics[width=\columnwidth]{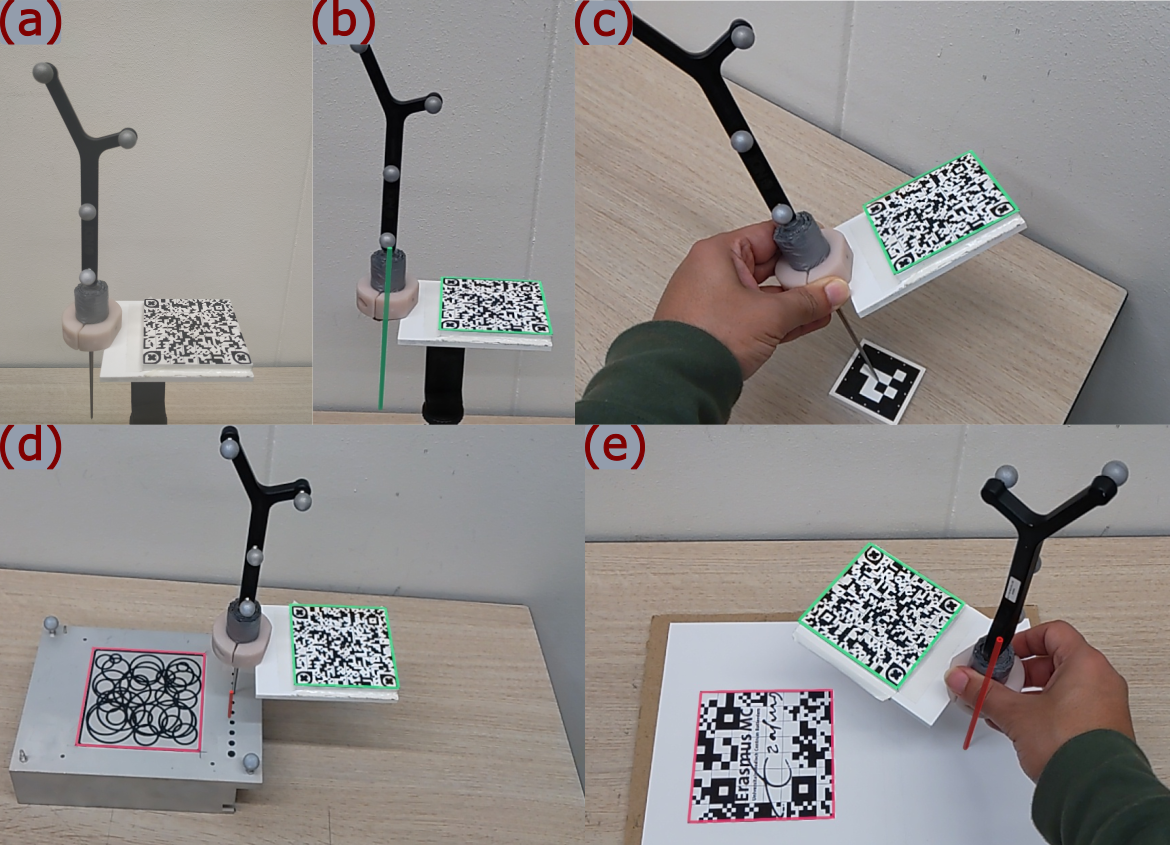}}
\caption{Tool-tip calibration experiment: (a) NDI pointer with a Vuforia marker attached. (b) The pointer tracked and calibrated in AR. (c-e) calibrating the pointer using pivot calibration, a calibration tool, and using marker-to-marker user calibration.}
\label{fig:exp_calib}
\end{figure}

\begin{figure}[]
\centerline{\includegraphics[width=\columnwidth]{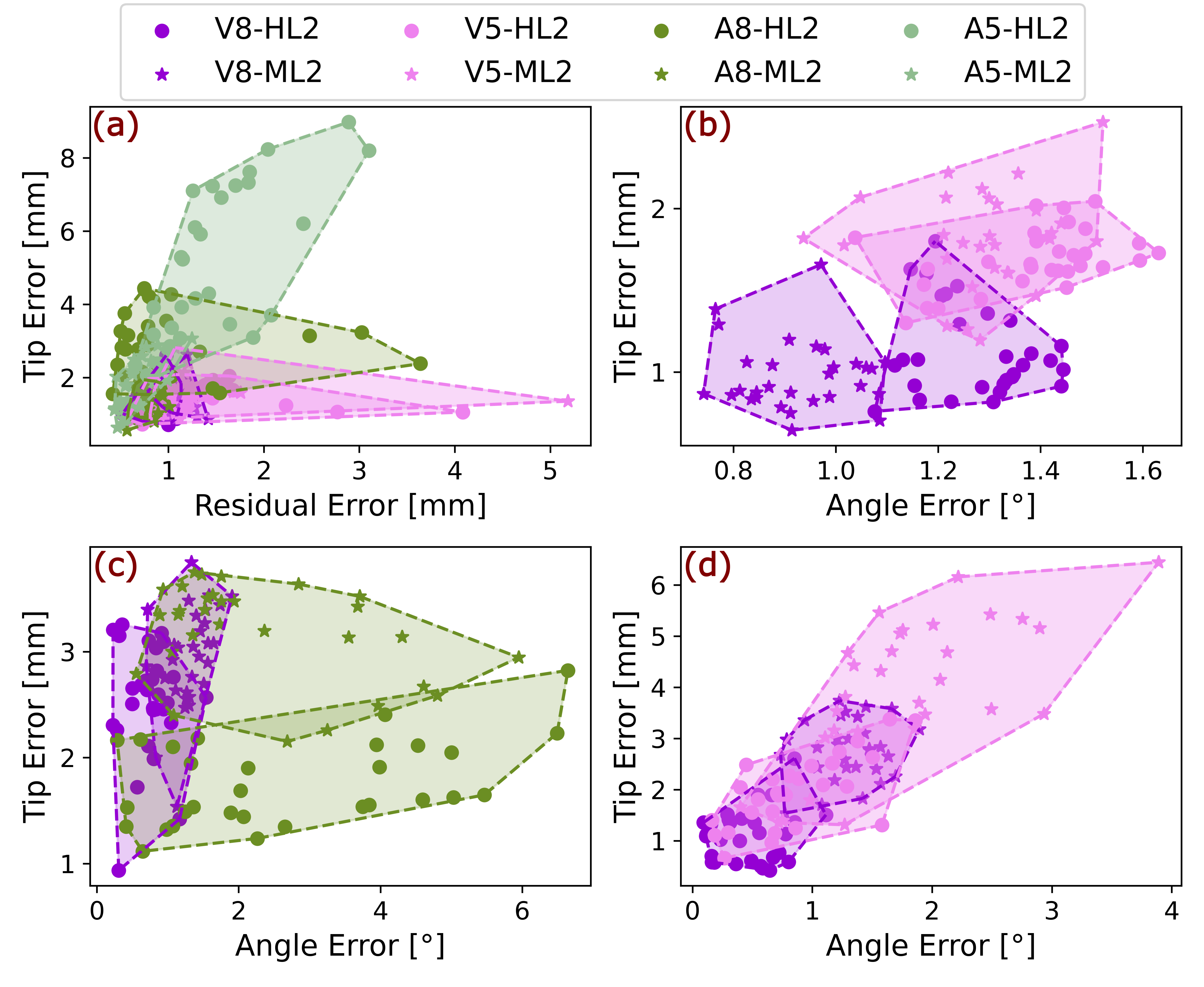}}
\caption{Tool-tip calibration error: (a) pivot calibration, (b) using a calibration tool, (c-d) marker-to-marker user calibration with reference marker size of 8 cm (c) and 12 cm (d). V8-HL2 in the legend indicates the marker type (Vuforia), marker size (8 cm) and HMD device (HL2).}
\label{fig:calib_exp_res}
\end{figure}

\begin{table*}[]
\centering
\caption{Tool-tip calibration error [mean(std)] of the proposed calibration approaches for different HMD devices and markers, with (P) pointer marker, (T) tool marker, and (R) reference marker.}
\label{tab:calib_exp}
\resizebox{\textwidth}{!}{%
\begin{tabular}{@{}cccccccccccccc@{}}
\toprule
\multirow{3}{*}{\textbf{Device}} & \multirow{3}{*}{\textbf{Marker}} & \multicolumn{2}{c}{\textbf{Pivot calibration}} & \multicolumn{4}{c}{\textbf{Tool calibration}} & \multicolumn{6}{c}{\textbf{Marker-to-marker user calibration}} \\ \cmidrule(l){3-14} 
 &  & 8 cm (P) & 5 cm (P) & \multicolumn{2}{c}{8 cm (P) - 10 cm (T)} & \multicolumn{2}{c}{5 cm (P) - 10 cm (T)} & \multicolumn{2}{c}{8 cm (P) - 8 cm (R)} & \multicolumn{2}{c}{8 cm (P) - 12 cm (R)} & \multicolumn{2}{c}{5 cm (P) - 12 cm (R)} \\
 &  & Tip (mm) & Tip (mm) & Tip (mm) & Angle (\degree) & Tip (mm) & Angle (\degree) & Tip (mm) & Angle (\degree) & Tip (mm) & Angle (\degree) & Tip (mm) & Angle (\degree) \\ \midrule
\multirow{2}{*}{HL2} & ArUco & 2.9 (0.9) & 4.9 (2.1) &  &  &  &  & \textbf{1.8 (0.4)} & \textbf{2.7 (1.9)} &  &  &  &  \\
 & Vuforia & \textbf{1.2 (0.4)} & \textbf{1.5 (0.4)} & \textbf{1.1 (0.3)} & \textbf{1.3 (0.1)} & \textbf{1.7 (0.2)} & \textbf{1.4 (0.1)} & 2.5 (0.5) & 0.7 (0.3) & \textbf{1.1 (0.5)} & \textbf{0.4 (0.3)} & \textbf{2.0 (0.7)} & \textbf{0.9 (0.4)} \\ \midrule
\multirow{2}{*}{ML2} & ArUco & 1.4 (0.3) & 1.8 (0.7) &  &  &  &  & 3.2 (0.4) & 2.3 (1.4) &  &  &  &  \\
 & Vuforia & \textbf{1.1 (0.4)} & \textbf{1.5 (0.4)} & \textbf{1.0 (0.2)} & \textbf{0.9 (0.1)} & \textbf{1.8 (0.3)} & \textbf{1.3 (0.1)} & 2.9 (0.5) & 1.3 (0.3) & 2.8 (0.6) & 1.3 (0.3) & 3.9 (1.4) & 1.8 (0.7) \\ \midrule
NDI &  & 0.66 (0.1) &  & 0.58 (0.0) & 1.0 (0.0) &  &  & 0.5 (0.1) & 1.2 (0.3) &  &  &  &  \\ \bottomrule
\end{tabular}%
}
\end{table*}

\subsection{Calibration of surgical instruments}
\label{sec:calibration_exp}

The calibration methods from Section~\ref{subsec: toolsCalib} were tested and evaluated using both ArUco and Vuforia markers. A mounting back-plate was rigidly attached to the NDI optical pointer (Figure \ref{fig:exp_calib}a), allowing the markers to be swapped while maintaining the same pose relative to the tool-tip.

To establish the ground truth tool-tip pose relative to the attached marker, the tool-tip was first determined in the NDI pointer's coordinates using the NDI software. The marker was then calibrated to the pointer by identifying the divots on the marker within the NDI pointer's coordinates. The calibration methods were evaluated by comparing the estimated tool-tip pose to the ground-truth pose.

During the experiment, three HL2 and one ML2 devices were available. The calibration tests for each method were repeated ten times for each HL2 device and 30 times for the ML2 device. At each evaluation test, both NDI and HMD tracking data were collected and processed for tool-tip pose estimation, which were evaluated against the ground truth tool tip poses established for both coordinate systems.

\begin{figure}[!b]
\centerline{\includegraphics[width=\columnwidth]{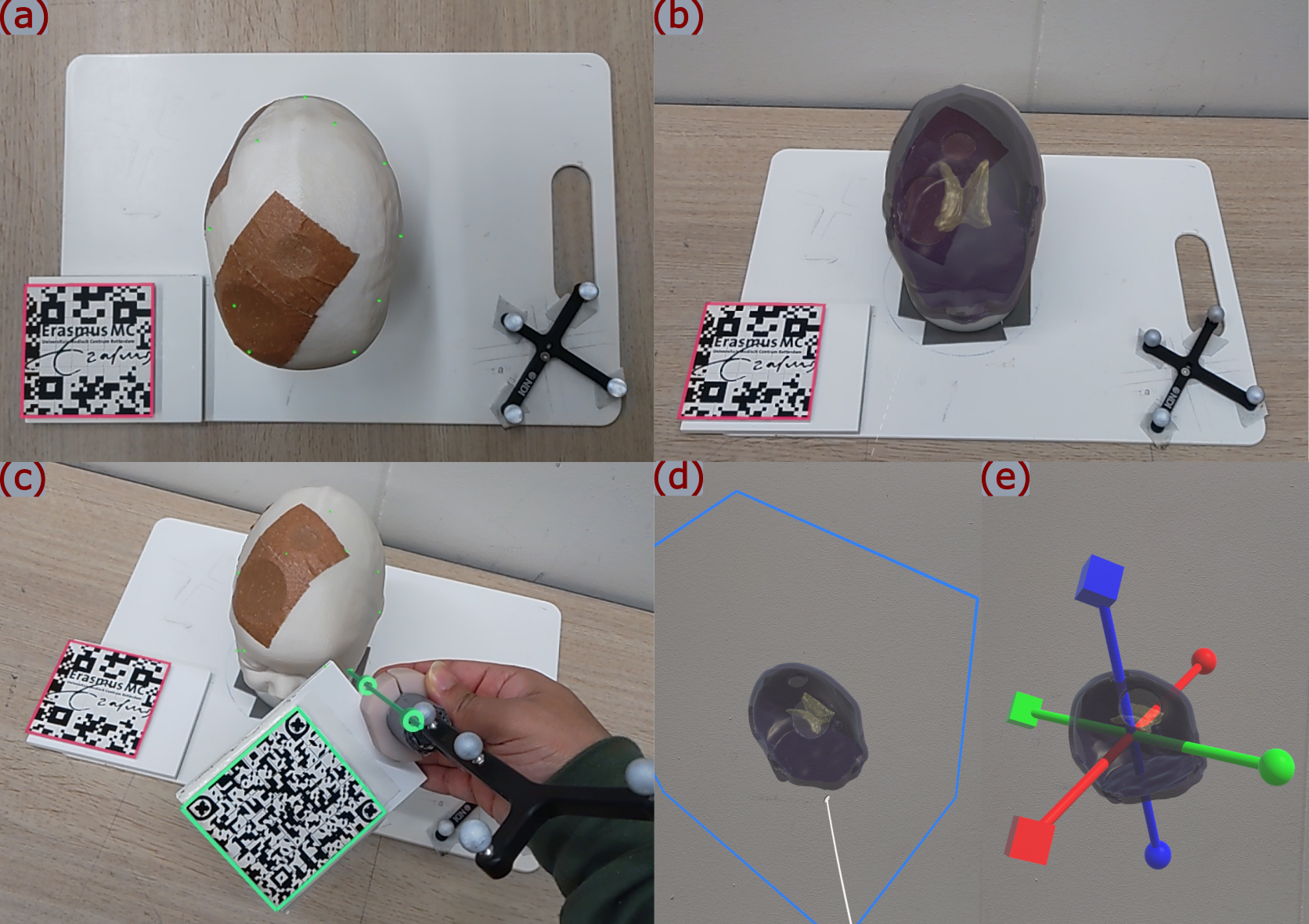}}
\caption{Image-to-patient alignment experiment: (a) patient-board with Vuforia and NDI markers attached, (b) aligned virtual model, (c) points annotation using the pointer, (d-e) manual positioning using 6-DoF and 1-DoF manipulation.}
\label{fig:reg_exp}
\end{figure}

\subsubsection{Pivot calibration}
\label{sec:pivot_calib_exp}
In this experiment, the pointer with the attached Vuforia/ArUco marker, tracked by both the NDI and HMD, was placed on a divot and pivoted slowly, ensuring the marker remained visible in the HMD camera (Figure \ref{fig:exp_calib}c). Marker poses were recorded for 40 seconds, and the tool-tip pose for both NDI and HMD was calculated using Equation \ref{eq:pivot}.

\begin{table*}[]
\centering
\caption{HL2 registration error {[}mean(std){]} for point-based matching for different combinations of points and markers.}
\label{tab:pb_reg_res_hl2}
\resizebox{\textwidth}{!}{%
\begin{tabular}{@{}ccccccccccc@{}}
\toprule
\multirow{4}{*}{\textbf{\begin{tabular}[c]{@{}c@{}}Number of \\ registration \\ points\end{tabular}}} & \multirow{4}{*}{\textbf{Annotation}} & \multirow{4}{*}{\textbf{Error metric}} & \multicolumn{4}{c}{\multirow{2}{*}{\textbf{Vuforia}}} & \multicolumn{4}{c}{\multirow{2}{*}{\textbf{ArUco}}} \\
 &  &  & \multicolumn{4}{c}{} & \multicolumn{4}{c}{} \\ \cmidrule(l){4-11} 
 &  &  & \multicolumn{2}{c}{\textbf{8 cm}} & \multicolumn{2}{c}{\textbf{5 cm}} & \multicolumn{2}{c}{\textbf{8 cm}} & \multicolumn{2}{c}{\textbf{5 cm}} \\
 &  &  & \textbf{GT-ttp} & \textbf{Pivot-ttp} & \textbf{GT-ttp} & \textbf{Pivot-ttp} & \textbf{GT-ttp} & \textbf{Pivot-ttp} & \textbf{GT-ttp} & \textbf{Pivot-ttp} \\ \midrule
\multirow{5}{*}{10 points} & \multirow{3}{*}{\begin{tabular}[c]{@{}c@{}}Landmarks\\ (mm)\end{tabular}} & NDI FRE & \multicolumn{2}{c}{1.2 (0.1)} & \multicolumn{2}{c}{1.1 (0.0)} & \multicolumn{2}{c}{1.1 (0.1)} & \multicolumn{2}{c}{1.1 (0.1)} \\
 &  & HMD FRE & 2.2 (0.3) & 2.8 (0.3) & 2.2 (0.3) & 2.7 (0.3) & 1.5 (0.3) & 1.6 (0.3) & 2.1 (0.3) & 2.1 (0.4) \\
 &  & HMD TRE (NDI-10) & \textbf{2.1 (0.3)} & \textbf{2.6 (0.3)} & \textbf{2.7 (0.5)} & \textbf{2.4 (0.5)} & \textbf{3.0 (0.3)} & \textbf{3.2 (0.3)} & \textbf{3.6 (0.3)} & \textbf{4.1 (0.4)} \\
 & \multirow{2}{*}{Trajectories} & Distance (mm) & 1.6 (0.4) & 2.2 (0.4) & 2.0 (0.6) & 2.0 (0.5) & 2.6 (0.3) & 2.8 (0.4) & 3.4 (0.3) & 3.5 (0.4) \\
 &  & Angle (\degree) & 0.9 (0.4) & 0.9 (0.4) & 1.4 (0.3) & 0.9 (0.3) & 1.5 (0.4) & 3.1 (0.5) & 1.5 (0.4) & 3.7 (0.6) \\ \midrule
\multirow{6}{*}{5 points} & \multirow{4}{*}{\begin{tabular}[c]{@{}c@{}}Landmarks\\ (mm)\end{tabular}} & NDI FRE & \multicolumn{2}{c}{1.1 (0.0)} & \multicolumn{2}{c}{1.1 (0.0)} & \multicolumn{2}{c}{1.0 (0.2)} & \multicolumn{2}{c}{1.1 (0.2)} \\
 &  & NDI TRE (5) & \multicolumn{2}{c}{1.5 (0.1)} & \multicolumn{2}{c}{1.4 (0.1)} & \multicolumn{2}{c}{1.6 (0.1)} & \multicolumn{2}{c}{1.5 (0.2)} \\
 &  & HMD FRE & 2.0 (0.2) & 2.6 (0.2) & 2.0 (0.4) & 2.7 (0.4) & 1.2 (0.4) & 1.2 (0.3) & 1.7 (0.5) & 1.7 (0.5) \\
 &  & HMD TRE (NDI-10) & \textbf{2.2 (0.3)} & \textbf{2.6 (0.3)} & \textbf{2.8 (0.5)} & \textbf{2.7 (0.5)} & \textbf{2.9 (0.4)} & \textbf{3.3 (0.3)} & \textbf{3.5 (0.4)} & \textbf{4.1 (0.3)} \\
 & \multirow{2}{*}{Trajectories} & Distance (mm) & 1.7 (0.5) & 2.2 (0.5) & 2.0 (0.6) & 2.4 (0.7) & 2.7 (0.3) & 3.4 (0.5) & 3.6 (0.5) & 3.9 (0.4) \\
 &  & Angle (\degree) & 1.1 (0.6) & 1.1 (0.6) & 1.7 (0.5) & 1.1 (0.5) & 1.3 (0.2) & 2.7 (0.4) & 1.5 (0.6) & 3.2 (0.7) \\ \midrule
\multirow{6}{*}{6 points} & \multirow{4}{*}{\begin{tabular}[c]{@{}c@{}}Landmarks\\ (mm)\end{tabular}} & NDI FRE & \multicolumn{2}{c}{0.8 (0.0)} & \multicolumn{2}{c}{0.8 (0.0)} & \multicolumn{2}{c}{0.7 (0.2)} & \multicolumn{2}{c}{0.8 (0.2)} \\
 &  & NDI TRE (4) & \multicolumn{2}{c}{2.2 (0.1)} & \multicolumn{2}{c}{2.0 (0.1)} & \multicolumn{2}{c}{2.1 (0.1)} & \multicolumn{2}{c}{2.1 (0.1)} \\
 &  & HMD FRE & 1.4 (0.2) & 1.8 (0.3) & 1.7 (0.4) & 2.3 (0.4) & 1.2 (0.4) & 1.2 (0.4) & 2.0 (0.4) & 1.9 (0.5) \\
 &  & HMD TRE (NDI-10) & \textbf{2.8 (0.5)} & \textbf{3.4 (0.6)} & \textbf{3.8 (0.4)} & \textbf{3.8 (0.4)} & \textbf{3.2 (0.3)} & \textbf{3.5 (0.2)} & \textbf{4.0 (0.4)} & \textbf{4.7 (0.5)} \\
 & \multirow{2}{*}{Trajectories} & Distance (mm) & 2.3 (0.8) & 2.9 (0.9) & 2.9 (0.7) & 3.5 (0.7) & 2.5 (0.3) & 2.5 (0.3) & 3.6 (0.3) & 3.6 (0.6) \\
 &  & Angle (\degree) & 1.5 (0.5) & 1.6 (0.5) & 2.2 (0.4) & 2.0 (0.5) & 1.6 (0.3) & 3.2 (0.4) & 2.2 (0.7) & 4.2 (0.8) \\ \bottomrule
\end{tabular}%
}
\end{table*}

\begin{table*}[]
\centering
\caption{ML2 registration error {[}mean(std){]} for point-based matching for different combinations of points and markers.}
\label{tab:pb_reg_res_ml2}
\resizebox{\textwidth}{!}{%
\begin{tabular}{@{}ccccccccccc@{}}
\toprule
\multirow{4}{*}{\textbf{\begin{tabular}[c]{@{}c@{}}Number of \\ registration \\ points\end{tabular}}} & \multirow{4}{*}{\textbf{Annotation}} & \multirow{4}{*}{\textbf{Error metric}} & \multicolumn{4}{c}{\multirow{2}{*}{\textbf{Vuforia}}} & \multicolumn{4}{c}{\multirow{2}{*}{\textbf{ArUco}}} \\
 &  &  & \multicolumn{4}{c}{} & \multicolumn{4}{c}{} \\ \cmidrule(l){4-11} 
 &  &  & \multicolumn{2}{c}{\textbf{8 cm}} & \multicolumn{2}{c}{\textbf{5 cm}} & \multicolumn{2}{c}{\textbf{8 cm}} & \multicolumn{2}{c}{\textbf{5 cm}} \\
 &  &  & \textbf{GT-ttp} & \textbf{Pivot-ttp} & \textbf{GT-ttp} & \textbf{Pivot-ttp} & \textbf{GT-ttp} & \textbf{Pivot-ttp} & \textbf{GT-ttp} & \textbf{Pivot-ttp} \\ \midrule
\multirow{5}{*}{10 points} & \multirow{3}{*}{\begin{tabular}[c]{@{}c@{}}Landmarks\\ (mm)\end{tabular}} & NDI FRE & \multicolumn{2}{c}{1.2 (0.0)} & \multicolumn{2}{c}{1.2 (0.0)} & \multicolumn{2}{c}{1.2 (0.0)} & \multicolumn{2}{c}{1.2 (0.0)} \\
 &  & HMD FRE & 2.3 (0.3) & 2.4 (0.3) & 2.1 (0.3) & 2.7 (0.4) & 1.8 (0.2) & 1.9 (0.2) & 2.9 (0.5) & 2.2 (0.5) \\
 &  & HMD TRE (NDI-10) & \textbf{2.1 (0.2)} & \textbf{2.7 (0.2)} & \textbf{2.0 (0.3)} & \textbf{2.0 (0.3)} & \textbf{2.4 (0.2)} & \textbf{2.6 (0.3)} & \textbf{3.0 (0.3)} & \textbf{2.2 (0.3)} \\
 & \multirow{2}{*}{Trajectories} & Distance (mm) & 1.3 (0.4) & 1.9 (0.4) & 1.1 (0.4) & 1.4 (0.4) & 1.6 (0.1) & 1.8 (0.4) & 2.4 (0.3) & 1.5 (0.2) \\
 &  & Angle (\degree) & 1.5 (0.4) & 2.6 (0.4) & 1.5 (0.5) & 1.3 (0.5) & 1.7 (0.4) & 2.5 (0.4) & 1.7 (0.3) & 0.9 (0.4) \\ \midrule
\multirow{6}{*}{5 points} & \multirow{4}{*}{\begin{tabular}[c]{@{}c@{}}Landmarks\\ (mm)\end{tabular}} & NDI FRE & \multicolumn{2}{c}{1.1 (0.0)} & \multicolumn{2}{c}{1.1 (0.0)} & \multicolumn{2}{c}{1.0 (0.0)} & \multicolumn{2}{c}{1.0 (0.0)} \\
 &  & NDI TRE (5) & \multicolumn{2}{c}{1.6 (0.1)} & \multicolumn{2}{c}{1.4 (0.1)} & \multicolumn{2}{c}{1.5 (0.1)} & \multicolumn{2}{c}{1.6 (0.1)} \\
 &  & HMD FRE & 2.4 (0.4) & 2.5 (0.5) & 2.1 (0.4) & 2.8 (0.5) & 1.7 (0.3) & 1.6 (0.3) & 2.7 (0.8) & 2.1 (0.8) \\
 &  & HMD TRE (NDI-10) & \textbf{2.6 (0.3)} & \textbf{3.0 (0.3)} & \textbf{2.5 (0.6)} & \textbf{2.4 (0.6)} & \textbf{3.5 (0.3)} & \textbf{3.6 (0.5)} & \textbf{4.0 (0.8)} & \textbf{3.1 (0.8)} \\
 & \multirow{2}{*}{Trajectories} & Distance (mm) & 2.0 (0.6) & 1.9 (0.7) & 1.3 (1.1) & 1.5 (1.0) & 2.5 (0.3) & 2.3 (0.6) & 3.1 (0.5) & 2.0 (0.4) \\
 &  & Angle (\degree) & 1.7 (0.5) & 2.8 (0.6) & 1.9 (0.7) & 1.6 (0.8) & 2.5 (0.7) & 3.2 (0.7) & 2.6 (0.7) & 1.8 (0.8) \\ \midrule
\multirow{6}{*}{6 points} & \multirow{4}{*}{\begin{tabular}[c]{@{}c@{}}Landmarks\\ (mm)\end{tabular}} & NDI FRE & \multicolumn{2}{c}{0.8 (0.0)} & \multicolumn{2}{c}{0.8 (0.0)} & \multicolumn{2}{c}{0.7 (0.0)} & \multicolumn{2}{c}{0.7 (0.0)} \\
 &  & NDI TRE (4) & \multicolumn{2}{c}{2.2 (0.1)} & \multicolumn{2}{c}{2.1 (0.1)} & \multicolumn{2}{c}{2.1 (0.1)} & \multicolumn{2}{c}{2.2 (0.1)} \\
 &  & HMD FRE & 1.9 (0.4) & 1.9 (0.4) & 2.0 (0.5) & 2.5 (0.5) & 1.5 (0.4) & 1.6 (0.4) & 1.9 (0.3) & 1.3 (0.3) \\
 &  & HMD TRE (NDI-10) & \textbf{3.0 (0.3)} & \textbf{3.6 (0.4)} & \textbf{3.0 (0.5)} & \textbf{3.3 (0.5)} & \textbf{2.5 (0.3)} & \textbf{3.1 (0.4)} & \textbf{2.2 (0.2)} & \textbf{1.8 (0.2)} \\
 & \multirow{2}{*}{Trajectories} & Distance (mm) & 1.8 (0.5) & 1.5 (0.5) & 1.7 (0.6) & 2.5 (0.6) & 1.9 (0.3) & 2.4 (0.5) & 1.8 (0.2) & 1.7 (0.3) \\
 &  & Angle (\degree) & 2.2 (0.5) & 3.3 (0.5) & 2.0 (0.6) & 2.0 (0.6) & 1.8 (0.4) & 2.5 (0.4) & 1.6 (0.3) & 0.9 (0.3) \\ \bottomrule
\end{tabular}%
}
\end{table*}

\begin{figure*}[]
\centerline{\includegraphics[width=\textwidth]{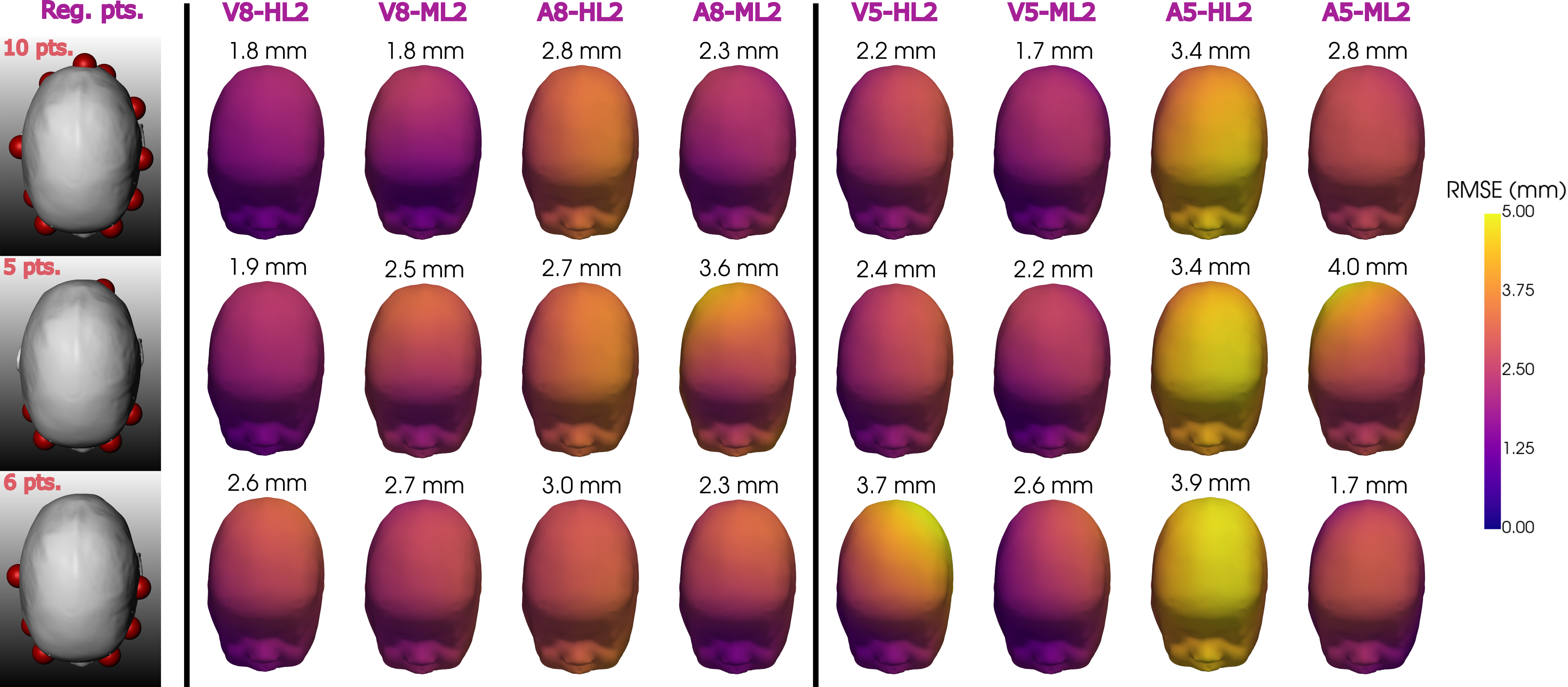}}
\caption{Surface-points mean registration error for the head phantom using 10, 5, and 6 registration points for both HL2 and ML2 using 8 and 5 cm marker sizes.}
\label{fig:res_pb_reg_surface}
\end{figure*}

\subsubsection{Using a calibration tool}
\label{sec:calib_tool_exp}
A calibration tool with a 10 cm Vuforia marker was designed. To determine the tool-tip pose, the pointer with the attached marker was inserted into a 3 mm diameter hole and kept static (Figure \ref{fig:exp_calib}d). The poses of the pointer and calibration tool were tracked by both NDI and HMD cameras. Marker poses were recorded for six seconds, and the tool-tip pose was calculated for both NDI and HMD using Equation \ref{eq:tool}.

\subsubsection{Marker-to-marker user calibration}
A static reference marker with a divot-point at a fixed distance of 10 cm or 15 cm was printed on paper. An NDI rigid body was mounted on the paper, and the divot's position relative to the rigid body was determined using the NDI pointer. The user then placed the pointer at the divot-point and visually aligned it with the virtual cylinder (Figure \ref{fig:exp_calib}e). The pointer and static reference marker poses were tracked by both NDI and HMD cameras. Marker poses were collected for six seconds, and the tool-tip pose was calculated for both NDI and HMD using Equation \ref{eq:user-calibration}.

Figure \ref{fig:calib_exp_res} and Table \ref{tab:calib_exp} show the tip error and angle deviation for each calibration method, marker type, marker size and HMD device. Tool calibration and marker-to-marker user calibration with different marker sizes were only evaluated for Vuforia markers. Table \ref{tab:calib_exp} also shows the calibration error of the pointer in NDI coordinate system for each calibration method. Tool-tip calibration with Vuforia markers reported lower error compared to calibration with ArUco markers for both HL2 and ML2. For the different methods, a tool-tip calibration error of around 1 mm distance and 1\degree~angle deviation was achieved using the framework.

\subsection{Image-to-patient alignment}

\subsubsection{Point-based matching accuracy assessment}
\label{sec:pb_reg_exp}
A head model from a CT scan was 3D printed with ten divots on its surface as landmarks. The head was fixated on a board with an NDI rigid body and a back-plate, on which Vuforia/ArUco markers were mounted and calibrated to the NDI rigid body (Figure \ref{fig:reg_exp}a). The NDI pointer with a Vuforia/ArUco marker (discussed in Section \ref{sec:calibration_exp}) was used to pinpoint the divots on the head model (Figure~\ref{fig:reg_exp}c). Both the HMD and NDI tracked the markers and rigid bodies respectively. The annotated divots were used to compute the image-to-patient registration transform for each coordinate system. The HMD registration transform was then compared to the ground truth transform of the NDI. The following evaluation metrics were used to assess the registration accuracy:

\begin{itemize}
    \item Fiducial registration error (FRE): is the root mean squared error (RMSE) of the fiducials used in the registration: 
    \begin{equation} \label{eq:registration_rmse}
    RMSE = \sqrt{(\sum^{N}_{n=1}|P_p -(R~.~P_m +T)|^2)/N} \quad ,
\end{equation}
    where $P_m$ and $P_p$ are the $N$ divot-points used in determining the registration transform on the virtual model and patient phantom respectively. $R$ and $T$ are the rotation and translation of the registration transform.
    \item Target registration error (TRE): same as FRE, but with $P_m$ and $P_p$ being the divot-points that were not used in determining the registration transform, on the virtual model and patient phantom respectively.
    \item Target registration error NDI-10 (TRE NDI-10): is the TRE where $P_p$ are the 10-NDI divot-points on the patient transformed to the Vuforia/ArUco marker coordinate system, and $R$ and $T$ corresponds to the HMD registration transform.
    \item Trajectories deviation: is the distance and angulation of five annotated trajectories on the patient model (representing ventricular needle trajectories, see section \ref{sec:evd_exp}) that were transformed using the NDI and HMD registration transforms in the Vuforia/ArUco marker coordinate system.
    \item Surface-points deviation: is the distance between corresponding model surface points that were transformed using the NDI and HMD registration transforms in the Vuforia/ArUco marker coordinate system.
\end{itemize}

The registration experiments were conducted for each marker type (Vuforia \& ArUco), marker size (8 cm \& 5 cm), and HMD device (HL2 \& ML2), with the evaluation tests being repeated five times for each of the three available HL2 devices and 15 times for the available ML2 device.

\begin{figure}[]
\centerline{\includegraphics[width=\columnwidth]{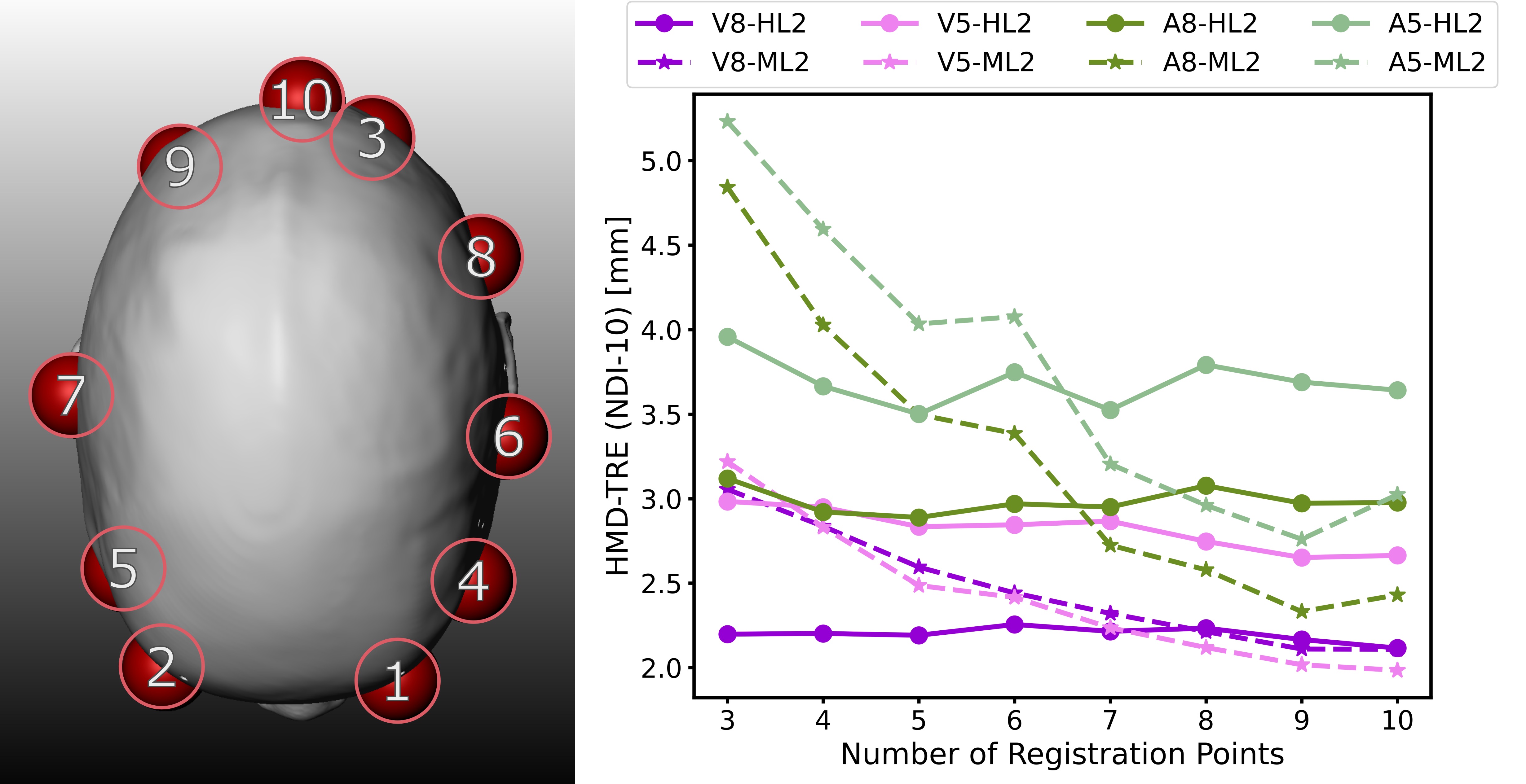}}
\caption{Registration error (HMD TRE NDI-10) for different number of reintegration points: (left) order of registration points, (right) registration error as the number of points increases from 3 to 10 points.}
\label{fig:reg_res_num_pts}
\end{figure}

Tables~\ref{tab:pb_reg_res_hl2} and~\ref{tab:pb_reg_res_ml2} show the mean registration error for both HL2 and ML2 devices, for each marker type and marker size using the ground truth tooltip of the pointer (GT-ttp) and the median estimated pivot tooltip from section \ref{sec:pivot_calib_exp} (Pivot-ttp). Three different arrangements of the registration points were evaluated. Figure \ref{fig:res_pb_reg_surface} shows the points arrangements as well as the surface-points deviation for each registration setup. The framework achieved on average an HMD TRE (NDI-10) of around 2-4.5 mm for the different configurations of markers, HMD devices, and registration points. The highest registration error was reported for ArUco 5 cm markers for HL2, while no substantial difference in registration error was observed for Vuforia markers between HL2 and ML2. For the arrangement of registration points, alignment with the 6-points arrangement reported the highest error compared to 10 and 5-points. Figure \ref{fig:reg_res_num_pts} further plots the registration error for the different setups for an increasing number of registration points. The figure shows a decrease in registration error for ML2 as the number of points increases, while error remains the same for HL2.

\begin{figure}[]
\centerline{\includegraphics[width=\columnwidth]{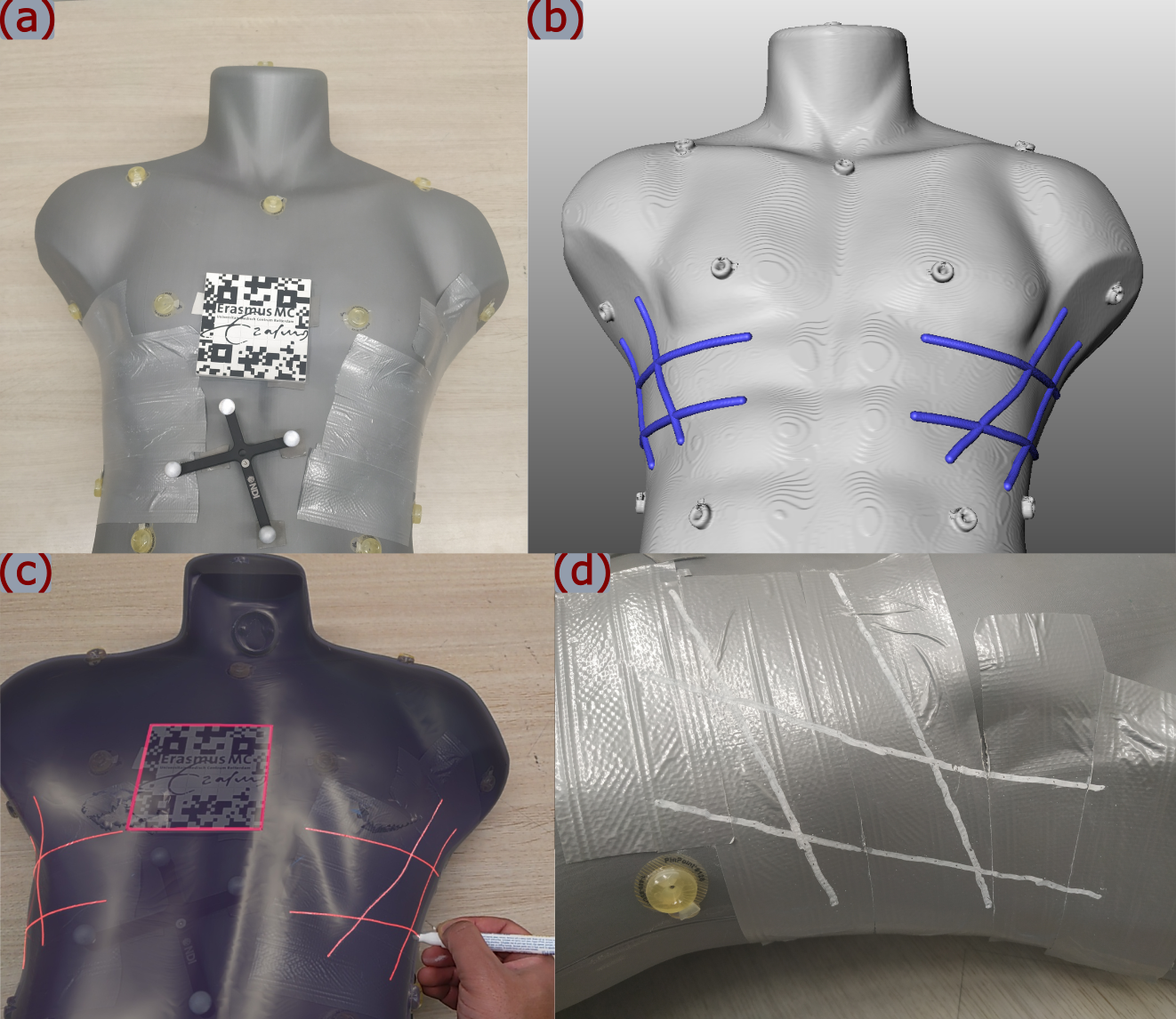}}
\caption{Locating rib-fractures experiment: (a) torso phantom with Vuforia and NDI markers, (b) planned incision lines, (c) aligned torso virtual model, (d) delineated incision lines by the user.}
\label{fig:rib_exp}
\end{figure}

\begin{figure}[]
\centerline{\includegraphics[width=\columnwidth]{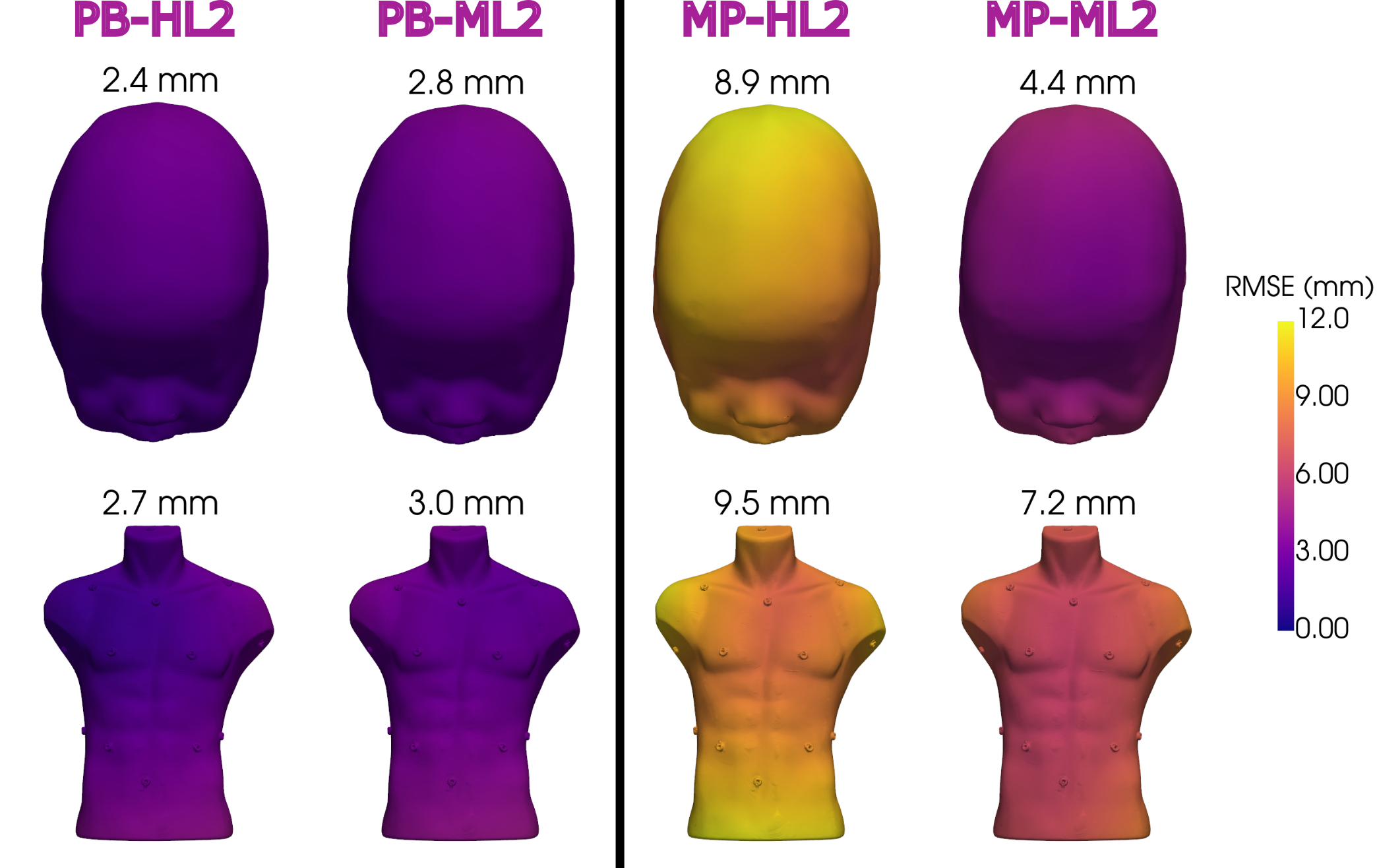}}
\caption{Surface-points mean registration error for point-based (PB) matching and manual positioning (MP) for both HL2 and ML2 performed by the participants on the head and torso phantoms.}
\label{fig:res_reg_manual}
\end{figure}

\begin{figure*}[]
\centerline{\includegraphics[width=\textwidth]{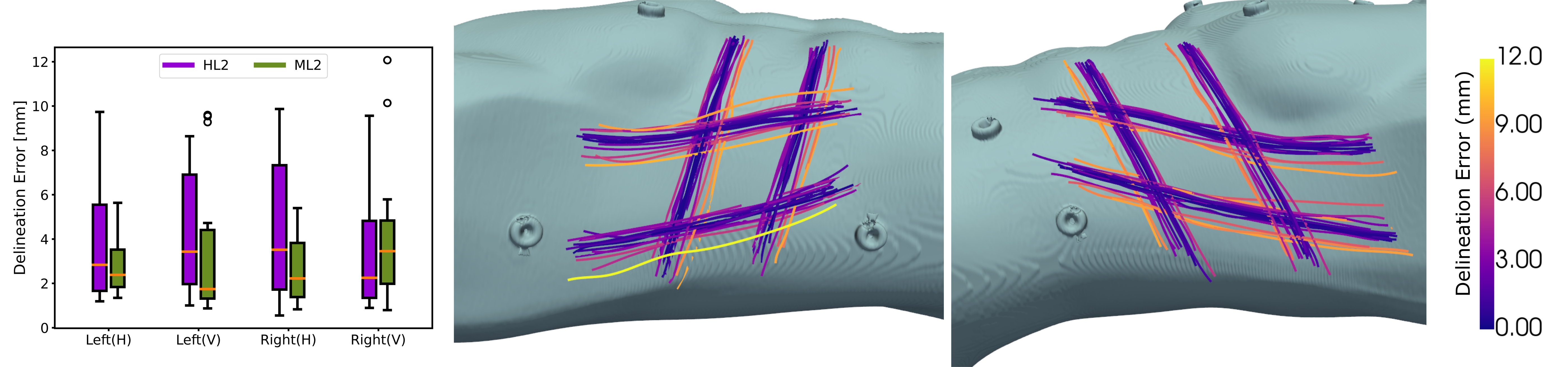}}
\caption{Delineation error of the incision lines by participants in the rib-fracture localization experiment: (left) Box plots of the delineation error for both HL2 and ML2. Left(H) and Left(V) represent the left-side horizontal and vertical planned incisions. (right) incision lines marked by the participants on both sides of the torso phantom.}
\label{fig:rib_results_incisions}
\end{figure*}

\begin{figure}[!b]
\centerline{\includegraphics[width=0.8\columnwidth]{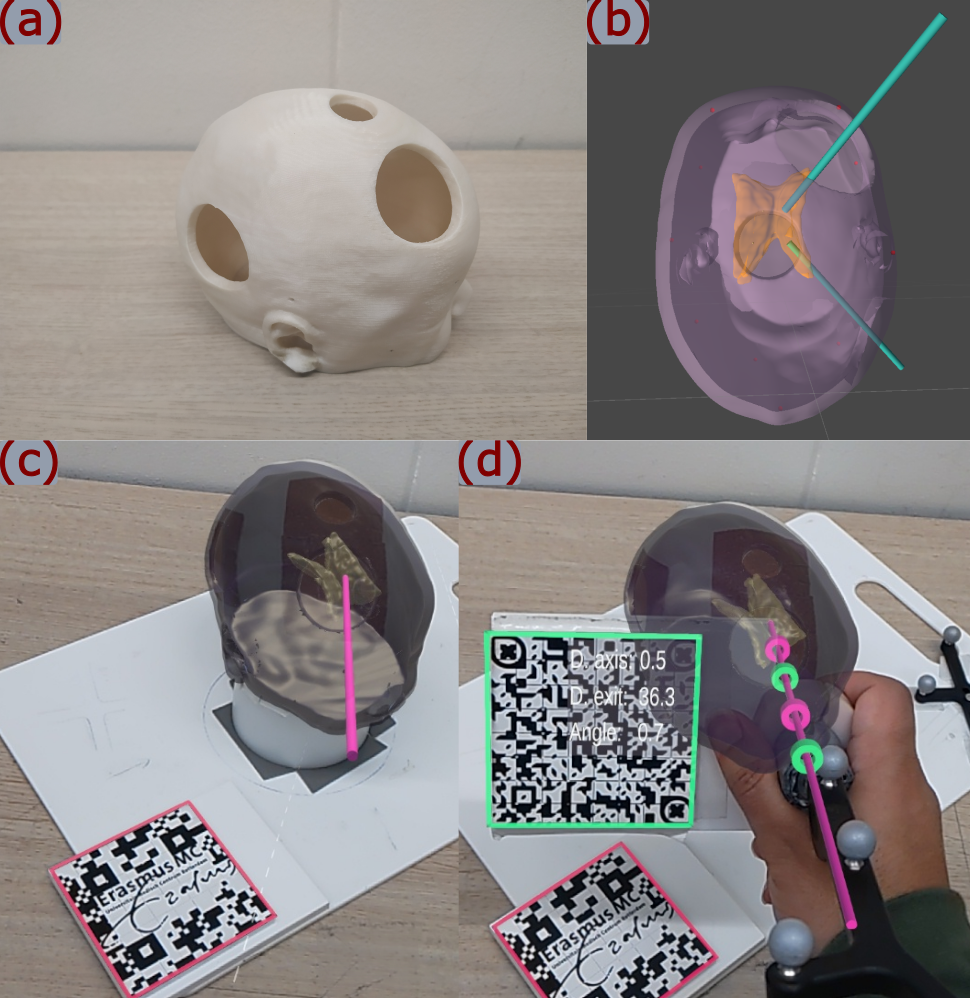}}
\caption{Ventricular catheter placement experiment: (a) head phantom with entry holes, (b) two insertion planned trajectories, (c) aligned virtual head with planned trajectory, (d) insertion guidance of the pointer with virtual extensions and numerical feedback menu.}
\label{fig:evd_exp}
\end{figure}

\subsubsection{Registration user experiment}
\label{sec:reg_user_exp}
 We evaluated point-based matching and manual-positioning for image-to-patient alignment in a user experiment with 16 participants who signed an informed consent to take part in the study. Participants were asked to align the preoperative model of a phantom using both registration approaches (Figure \ref{fig:reg_exp}). In addition to the head phantom, a torso phantom (Figure \ref{fig:rib_exp}a) was used. The torso phantom was CT scanned, and a 3D model was created for AR visualization. Twelve fiducial markers (PinPoint for Image Registration 128, Beekley) were placed on the torso for point-based matching. Participants were first shown a ground-truth alignment determined using the NDI system. They then performed the alignment task, starting with point-based matching (using six fiducial points) followed by manual positioning. The experiment used 8 cm Vuforia markers, with participants alternating between the HL2 and ML2 devices. The alignment approaches were evaluated in terms of accuracy (TRE-NDI and surface-point deviations) and completion time.

Table \ref{tab:res_reg_manual} and Figure \ref{fig:res_reg_manual} show the registration error for alignments performed by the participants for both manual-positioning and point-based matching on the two phantoms (head and torso). Point-based matching achieved lower registration error compared to manual-positioning for both HL2 and ML2 on both phantoms. Participants also took less time aligning the virtual model using point-based matching compared to manual-positioning. A higher difference in registration accuracy was observed between HL2 and ML2 for manual positioning compared to point-based matching.

\subsection{Targeting accuracy: locating rib-fractures}
\label{sec:rib_exp}
In trauma cases with subtle rib fractures or obese patients, locating the fracture can be challenging, potentially leading to suboptimal incisions with errors up to 5 cm \cite{thabit2024augmented}. This experiment evaluated the framework’s use in locating rib fractures in a phantom setup. Following the manual positioning of the torso model (described in Section \ref{sec:reg_user_exp}), participants were asked to delineate eight incision lines on the torso phantom using a 1 mm tip marker (Figure \ref{fig:rib_exp}). The lines were then digitized using the NDI system and transformed to the CT image space. The accuracy was assessed by calculating the distance between each drawn incision line and its corresponding model-annotated line.

The error of the delineated incisions after performing manual-positioning alignment of the torso model is shown in Figure \ref{fig:rib_results_incisions}. The mean delineation error was 3.9 mm for HL2 and 3.2 mm for ML2, with no significant difference between horizontal and vertical lines on the left and right sides of the torso phantom. Following the manual registration errors reported in section \ref{sec:reg_user_exp}, participants who used ML2 achieved a lower delineation error compared to participants who used HL2. The incision lines delineated by the participants for both HL2 and ML2 are shown in Figure \ref{fig:rib_results_incisions}.

\begin{table}[]
\centering
\caption{Point-based matching vs Manual-positioning in aligning head and torso virtual models. Accuracy is the HMD TRE NDI-10 [mean(std)] and time is the alignment completion time [mean(std)]}
\label{tab:res_reg_manual}
\resizebox{\columnwidth}{!}{%
\begin{tabular}{@{}llllll@{}}
\toprule
\multirow{2}{*}{\textbf{Device}} & \multirow{2}{*}{\textbf{Phantom}} & \multicolumn{2}{l}{\textbf{Point-based matching}} & \multicolumn{2}{l}{\textbf{Manual registration}} \\ \cmidrule(l){3-6} 
 &  & \textbf{Accuracy} & \textbf{Time} & \textbf{Accuracy} & \textbf{Time} \\ \midrule
\multirow{2}{*}{HL2 (n=8)} & Head & \textbf{2.6 (1.1)} & 2:38 (0:53) & 8.9 (2.8) & 8:25 (3:24) \\
 & Torso & \textbf{2.5 (1.0)} & 1:58 (1:04) & 9.4 (2.5) & 6:58 (3:08) \\ \midrule
\multirow{2}{*}{ML2 (n=8)} & Head & 3.0 (1.1) & 4:12 (2:21) & \textbf{4.6 (1.3)} & 4:57 (1:37) \\
 & Torso & 3.0 (1.0) & 3:01 (1:25) & \textbf{7.1 (2.2)} & 7:23 (4:24) \\ \midrule
\multirow{2}{*}{Total (n=16)} & Head & 2.8 (1.1) & 3:29 (1:59) & 6.7 (2.0) & 6:48 (3:13) \\
 & Torso & 2.8 (1.0) & 2:31 (1:22) & 8.3 (2.4) & 7:11 (3:49) \\ \bottomrule
\end{tabular}%
}
\end{table}

\subsection{Targeting accuracy: ventricular catheter placement}
\label{sec:evd_exp}
In neurosurgery, catheter placement for cerebrospinal fluid (CSF) drainage is typically done with a free hand approach, which may result in misplacement of the catheter~\cite{chai2013coma,alazri2017placement}. This experiment evaluates the framework's accuracy in guiding needle insertion for ventricular catheter placement. In our setup, two 5 cm diameter holes were made on a 3D-printed head phantom at Kocher's and Keen's points (common entry sites for catheter procedures). Five insertion trajectories were planned for each entry point (Figure \ref{fig:evd_exp}). A user experiment with 16 participants assessed the framework by guiding them to insert a pointer (simulating a needle) along the visualized trajectories. Prior to the experiment, the phantom was filled with an 8\% gelatin-water mixture to simulate brain tissue. The insertion holes were taped over, and participants were instructed to insert the pointer through the tape. Each participant performed 10 insertions, five from each entry point, alternating between the HL2 and ML2 devices.

For evaluation, the tool-tip calibration error, registration error, and insertion error were calculated. The insertion error is reported with the following metrics:
\begin{itemize}
    \item Perception error: The distance between the planned trajectory and the insertion tool in the HMD patient's coordinate system. This error represents user-introduced and perception errors.
    \item HMD error: The distance between the planned trajectory and the insertion tool in the HMD patient's coordinate system, using the ground truth tool-tip (as determined in Section~\ref{sec:calibration_exp}) and the registration based on ten divot-points. This error accounts for perception, calibration, and registration errors.
    \item NDI error: The distance between the planned trajectory and the insertion tool in the NDI patient's coordinate system. This error represents the total insertion error, encompassing perception, calibration, registration, and marker tracking errors for the HMD device.
\end{itemize}

The results of the user experiment for AR-guided needle insertion into the ventricle are shown in Table \ref{tab:evd_results}. The Table reports the tool-tip calibration error of the pointer using the calibration tool method (see section \ref{sec:calib_tool_exp}), as well as the registration error in the form of trajectories deviation. The Table also reports the needle insertion error, represented by the perception, HMD and NDI errors at the entry point, end point, and mean over insertion trajectory between entry and exit points. For both HL2 and ML2, participants achieved on average a perception error of 1-1.5 mm and 0.5-1\degree~at entry and exit points, while HMD and NDI insertion errors were around 2-3 mm and 2\degree~deviation. 

\subsection{Usability assessment}
\label{sec:usability_exp}

To evaluate the framework’s usability and user experience, participants in the user experiments on image-to-patient alignment, rib-fracture localization, and ventricular catheter placement (Sections~\ref{sec:reg_user_exp}--\ref{sec:evd_exp}) completed a feedback questionnaire. It assessed the perceived usefulness of the integrated functionalities and AR visualization for the tasks performed, as well as overall usability of the framework using the System Usability Scale (SUS) \cite{brooke1996sus} and NASA-TLX \cite{hart1988development} for surgical navigation.

A total of 16 participants took part in the user experiments and completed the feedback questionnaires. The participants had medical (37.5\%) and technical (62.5\%) backgrounds, with most participants (69\%) being unfamiliar with AR HMD devices. For image-to-patient registration, 75\% found point-based matching easier and more accurate than manual positioning for aligning virtual models. In manual positioning, the 1-DOF (54\%) and 6-DOF (46\%) object manipulation modes were considered comparably useful.

During the catheter placement task, participants generally observed good alignment between the virtual head model and the physical phantom. The augmented insertion tool was reported to follow the physical tool well, although occasional misalignments occurred due to incorrect orientation of the tool’s tracked marker. Both the virtual tool extensions (61\%) and the numeric feedback menu (39\%) were considered helpful in guiding the tool along the planned trajectory.

In terms of overall usability, the system achieved a mean SUS score of 74, indicating good usability. The NASA-TLX workload assessment yielded an average score of 40\%, with mental demand (51\%) and effort (56\%) contributing the most, while performance (26\%) and temporal demand (28\%) having lesser impact.

\begin{table*}[]
\centering
\caption{Mean calibration, registration and insertion errors for the ventricular catheter placement user study, for both HL2 and ML2. }
\label{tab:evd_results}
\resizebox{\textwidth}{!}{%
\begin{tabular}{@{}ccccccccccccc@{}}
\toprule
\multicolumn{2}{c}{\multirow{2}{*}{\textbf{Device}}} & \multirow{2}{*}{\textbf{\begin{tabular}[c]{@{}c@{}}Tool-tip calibration\\ error\end{tabular}}} & \multirow{2}{*}{\textbf{\begin{tabular}[c]{@{}c@{}}Registration error\\ (trajectories)\end{tabular}}} & \multicolumn{3}{c}{\textbf{Entry point}} & \multicolumn{3}{c}{\textbf{Exit point}} & \multicolumn{3}{c}{\textbf{Entry to exit (mean)}} \\ \cmidrule(l){5-13} 
\multicolumn{2}{c}{} &  &  & \textbf{Perception} & \textbf{HMD} & \textbf{NDI} & \textbf{Perception} & \textbf{HMD} & \textbf{NDI} & \textbf{Perception} & \textbf{HMD} & \textbf{NDI} \\ \midrule
\multirow{2}{*}{\begin{tabular}[c]{@{}c@{}}HL2\\ (n=8)\end{tabular}} & Distance (mm) & \textbf{1.2 (0.2)} & 2.6 (0.2) & 0.9 (0.6) & 3.2 (1.2) & 3.1 (1.6) & 1.4 (0.9) & 2.9 (0.9) & 3.3 (1.3) & 2.3 (2.3) & 3.5 (2.4) & 3 (1.0) \\
 & Angle (\degree) & \textbf{0.9 (0.3)} & 1.6 (0.2) & 0.7 (0.5) & 2.3 (0.7) & 2.3 (0.7) & 0.7 (0.3) & 2.3 (0.5) & 2.5 (0.7) & 1.3 (1.4) & 2.8 (1.3) & 2.7 (0.8) \\ \midrule
\multirow{2}{*}{\begin{tabular}[c]{@{}c@{}}ML2\\ (n=8)\end{tabular}} & Distance (mm) & 1.8 (0.7) & \textbf{2.2 (0.1)} & 0.9 (0.5) & 1.7 (0.7) & \textbf{2.0 (1.0)} & 1.4 (0.6) & 2.9 (0.8) & \textbf{2.6 (1.0)} & 2.3 (1.6) & 2.7 (1.6) & \textbf{2.2 (0.9)} \\
 & Angle (\degree) & 0.9 (0.8) & \textbf{0.7 (0.1)} & 0.6 (0.4) & 1.6 (0.7) & \textbf{1.8 (0.7)} & 0.8 (0.5) & 1.6 (0.7) & \textbf{1.7 (0.7)} & 1.5 (1.0) & 2.2 (1.0) & \textbf{2.2 (1.0)} \\ \midrule
\multirow{2}{*}{\begin{tabular}[c]{@{}c@{}}Total\\ (n=16)\end{tabular}} & Distance (mm) & 1.5 (0.6) & 2.4 (0.1) & 0.9 (0.6) & 2.4 (0.9) & 2.5 (1.3) & 1.4 (0.7) & 2.9 (0.8) & 3.0 (1.2) & 2.3 (2.0) & 3.1 (2.0) & 2.6 (0.9) \\
 & Angle (\degree) & 0.9 (0.6) & 1.2 (0.2) & 0.6 (0.4) & 1.9 (0.7) & 2.0 (0.7) & 0.7 (0.4) & 1.9 (0.6) & 2.1 (0.7) & 1.4 (1.2) & 2.5 (1.2) & 2.4 (0.9) \\ \bottomrule
\end{tabular}%
}
\end{table*}

\section{Discussion}

In this study, a framework for stand-alone surgical navigation using HMD devices was developed and evaluated. The framework supports the tracking of reference markers, calibration of surgical instruments, and the alignment and visualization of the preoperative data with the patient. The extensive evaluation demonstrated that the framework can achieve accuracies of 1 mm and 1\degree~in calibrating surgical instruments, 2-3 mm in image-to-patient alignment, and 2-4 mm in performing surgical tasks such as needle insertion and incision line localization for rib-fractures. This level of accuracy can be sufficient for a range of surgical applications, but not for procedures requiring sub-millimeter precision. 

The usability assessment of the framework showed that users found the various components of the framework useful and helpful in assisting the navigation tasks, such as the 1-DOF and 6-DOF object manipulation for manual alignment, and the virtual tool extensions and numeric feedback menu for needle insertion. Although most participants (69\%) were not familiar with AR HMD devices, they generally found the framework intuitive and easy to work with, with some noting that additional training time would improve their proficiency with the system.

The proposed framework integrates multiple methods for marker tracking, surgical tool calibration, and image-to-patient alignment. In marker tracking Vuforia demonstrated higher accuracy than ArUco in both instrument calibration and patient registration across the HL2 and ML2 devices. While Vuforia’s performance was consistent across the two devices, ArUco showed improved accuracy on the ML2, thanks to its built-in marker tracking extension.

For surgical tool calibration, the use of a dedicated calibration tool produced the most accurate results. However, this approach requires the construction of an additional instrument. In this study, a calibration tool was created using only a Vuforia marker, though an ArUco marker can be employed in a similar manner. Alternatively, pivot calibration or marker-to-marker user calibration can be used to determine the tool tip. The latter method relies on the user’s visual alignment of the tool’s orientation.

For image-to-patient alignment, point-based matching achieved higher accuracy than manual positioning and was less dependent on the user's skill and experience. Nevertheless, manual positioning remains valuable in cases where clear anatomical landmarks are absent, such as in the torso region, or when marker-less registration is required.

The user experiments for needle insertion and rib-fracture localization showed that participants were able to effectively follow the framework's AR guidance to perform common surgical tasks. Guiding surgical tools along planned trajectories is frequently required in procedures such as pedicle screw placement in spine surgery \cite{liebmann2019pedicle} and implant placement in cranio-maxillofacial surgery \cite{wu2019real}, whereas locating target structures and defining incision lines is of great value in neurosurgery and trauma applications~\cite{thabit2022augmented}. The framework can be readily adapted to such surgical applications via the configuration manager, enabling less technically experienced users to utilize AR guidance and supporting further validation in clinical settings.

Since the primary objective of this work was to propose a generalizable framework and demonstrate its usability across select surgical applications, the validation of the framework was performed in phantom setups, and therefore the reported accuracy mainly reflects the potential of applying the framework in clinical settings. To this end, the evaluation experiments focused on assessing the accuracy of the framework in relation to system-internal factors such as marker tracking, tooltip calibration and point-based matching. External, clinically dependent factors such as the positioning of markers, choice of anatomical landmarks, tissue movement and respiration were not considered in this assessment, as they vary across applications. Nevertheless, standard practices used in conventional navigation systems to mitigate errors arising from these clinical factors should also be applied when using the proposed framework. For instance, the patient marker should be positioned as close as possible to the target structure, and fiducial or anatomical landmarks should be well identified and distributed around the target object. Similar to conventional navigation systems, the framework cannot compensate for tissue deformation or when the patient anatomy differs substantially from the preoperative images. The impact of these clinical factors can be further investigated for specific surgical applications in future work.

The reported accuracy of the proposed framework (2-4 mm targeting error) is comparable to that of state-of-the-art HMD-based systems in the literature. For example, Liebman et al. reported a mean insertion error of 2.77 mm and 3.38° in pedicle screw placement for spine surgery, while their surface registration approach was specifically designed for vertebrae alignment~\cite{liebmann2019pedicle}. Martin-Gomez et al. reported an accuracy of 3.04 mm and 4.70° for guided k-wire insertion in orthopedic surgery; however, their system required a connection to a workstation for instrument tracking~\cite{martin2023sttar}. Benmahdjoub et al. evaluated their system for guiding incision placement in craniosynostosis surgery and reported an error of 3 mm~\cite{thabit2022augmented}, while Iqbal et al. reported a mean error of 2.03 mm and 1.12° for k-wire insertion in orthopedic surgery~\cite{iqbal2022semi}. Both systems, relied on connections to external navigation systems or robotic platforms. Gsaxner et al. reported an error of 4.93 mm in needle placement for tumor biopsy with on-device tracking with an additional mounted light source~\cite{gsaxner2021inside}. Li et al. achieved an error of 2.05 mm and 3.39° in simulated EVD injections with marker-less, depth-based registration, with the user manually initiating the alignment from a nearby workstation~\cite{li2024evd}.

In comparison, the proposed framework in this study is a stand-alone system that requires only the HMD device for navigation. It is generalizable, application-agnostic and device-agnostic. Although direct comparisons between the proposed framework and other systems are not possible due to the differences in experimental setup and surgical applications, making the framework publicly available can facilitate benchmarking of HMD-based AR surgical navigation systems. We hope providing such a generalizable and adaptable tool can support further development and evaluation in clinical settings, moving beyond initial phantom setups.

\section{Conclusion}
The proposed HMD-based surgical navigation framework integrates various methods for tracking reference markers, calibrating surgical instruments, aligning preoperative models with the patient and visualizing target structures. The modules of the framework allow it to be configurable and generalizable to multiple surgical applications. An extensive validation of the framework provided a benchmark for its accuracy and was assessed on the Microsoft HoloLens 2 and Magic Leap 2. The framework showed that it can achieve less than 5 mm targeting error in ventricular catheter placement and rib-fracture localization using the two HMD devices. The framework is publicly available and can be extended to integrate additional tracking and registration approaches. Furthermore, it can be used as a benchmark for other AR guidance techniques developed for HMD devices.

\bibliographystyle{IEEEtran}
\bibliography{bibliography-framework}

\end{document}